\documentclass[11pt]{article}
\usepackage{amsmath}

\usepackage[final]{acl}

\usepackage{times}
\usepackage{latexsym}

\usepackage{booktabs}
\usepackage{multirow}
\usepackage{amssymb} 
\usepackage{pifont}  
\usepackage{pgfplots}
\usepackage{pgfplotstable}
\usepgfplotslibrary{groupplots}
\usetikzlibrary{patterns}
\pgfplotsset{compat=1.17}

\usepackage{enumitem}
\usepackage{tcolorbox}
\tcbuselibrary{breakable}

\newtcolorbox[auto counter]{promptbox}[2][]{%
    colback=gray!5!white,
    colframe=gray!50!black,
    fonttitle=\bfseries,
    title={Prompt \thetcbcounter: #2}, 
    label={#1},
    breakable,
    arc=2mm,
    boxrule=0.5pt,
    fontupper=\small\ttfamily
}

\usepackage{array}

\usepackage{xcolor}
\usepackage[table]{xcolor}

\usepackage[T1]{fontenc}

\usepackage[utf8]{inputenc}

\usepackage{microtype}

\usepackage{inconsolata}

\usepackage{graphicx}

\title{OBJECTION! \\ Lawyer Agents Mitigate Guilty Bias in Legal Judgment Prediction}

\author{
  Jaehoon Jeong$^{1,2}$ \qquad Jay-Yoon Lee$^{1,}$\thanks{\ \ Corresponding author.} \\
  $^1$Seoul National University, Seoul, South Korea \\
  $^2$Korean National Police Agency, Seoul, South Korea \\
  \texttt{\{20210042, lee.jayyoon\}@snu.ac.kr}
}

\begin{document}

\maketitle
\begin{abstract}
Legal Judgment Prediction (LJP) models are typically trained on documents that describe facts from a prosecutorial perspective. Existing datasets further exhibit severe label imbalance toward guilty outcomes. Consequently, these models suffer from ``Guilty Bias'', blindly accepting the prosecution’s narrative as objective truth. Previous studies employing three-step reasoning structures or training on synthetically generated innocence data improve overall accuracy, but they still fail to mitigate bias at inference time.

In this paper, we introduce OBJECTION, an inference-time pipeline that integrates an Adversarial Lawyer Agent into each 3-step reasoning of offense, unlawfulness, and culpability. Unlike generic critics, our agent actively challenges the model’s presumptions of guilt by injecting legal defense arguments at each reasoning stage. To thoroughly evaluate this, we present a new ``Natural Innocent'' dataset including 3.4k real-world cases\footnote{Available at \url{https://huggingface.co/datasets/Kcsp0042/natural-innocent}. The public release (3,200 cases) is a revised edition of the data used here.}, overcoming the limitations of synthetic innocence benchmarks. Test results show that OBJECTION drastically reduces the False Guilty Rate (FGR) from 82.93\% (SOTA baseline) to 16.69\%, proving its capability to perform substantive legal reasoning. This work denotes a key progress toward aligning Legal AI with the presumption of innocence.

\end{abstract}

\section{Introduction}
The maxim “It is better that ten guilty persons escape than that one innocent suffer” epitomizes the presumption of innocence, a basic tenet of criminal law. However, criminal trials are not neutral by design. The very fact that a case has reached the courtroom implies that investigative bodies have established a sufficient suspicion of guilt. Consequently, the ‘written judgment’ inevitably reflects this bias. These fact descriptions are not neutral event summaries, but prosecutorial narratives written to justify guilt.

The critical issue is that the vast majority of legal texts used to train LLMs consist of precisely this type of content. In Legal Judgment Prediction (LJP), models are typically trained on these prosecutions' stated facts \cite{luo2017learning, xiao2018cail, zhong2018legal, cui2023survey}. As a result, models are systematically exposed to guilt-biased narratives even if the actual verdict is not guilty. This guilt-biased nature poses a serious impediment to real-world application. Convicting an innocent individual undermines public trust; thus, a high False Guilty Rate (FGR) is impractical and normatively unacceptable.

Previous research has attempted to mitigate this bias by structured reasoning, such as explicit element schemas \cite{liu2025jurex} or multi-step legal reasoning frameworks \cite{jiang2024legal, deng2023syllogistic, zhang2025syler}. However, these approaches fundamentally assume that the input fact description is unbiased. As a result, they inherit the guilty bias embedded in the narrative itself. Even recent efforts based on synthetic innocent cases \cite {zhang2025ljpiv} largely rely on training-time correction. Such synthetic innocence often introduces artificial cues that models can learn as shortcuts, leading to overfitting, which we empirically observe, while inference-time bias remains unaddressed.

\begin{figure*}[ht]
  \centering
  \includegraphics[width= \linewidth]{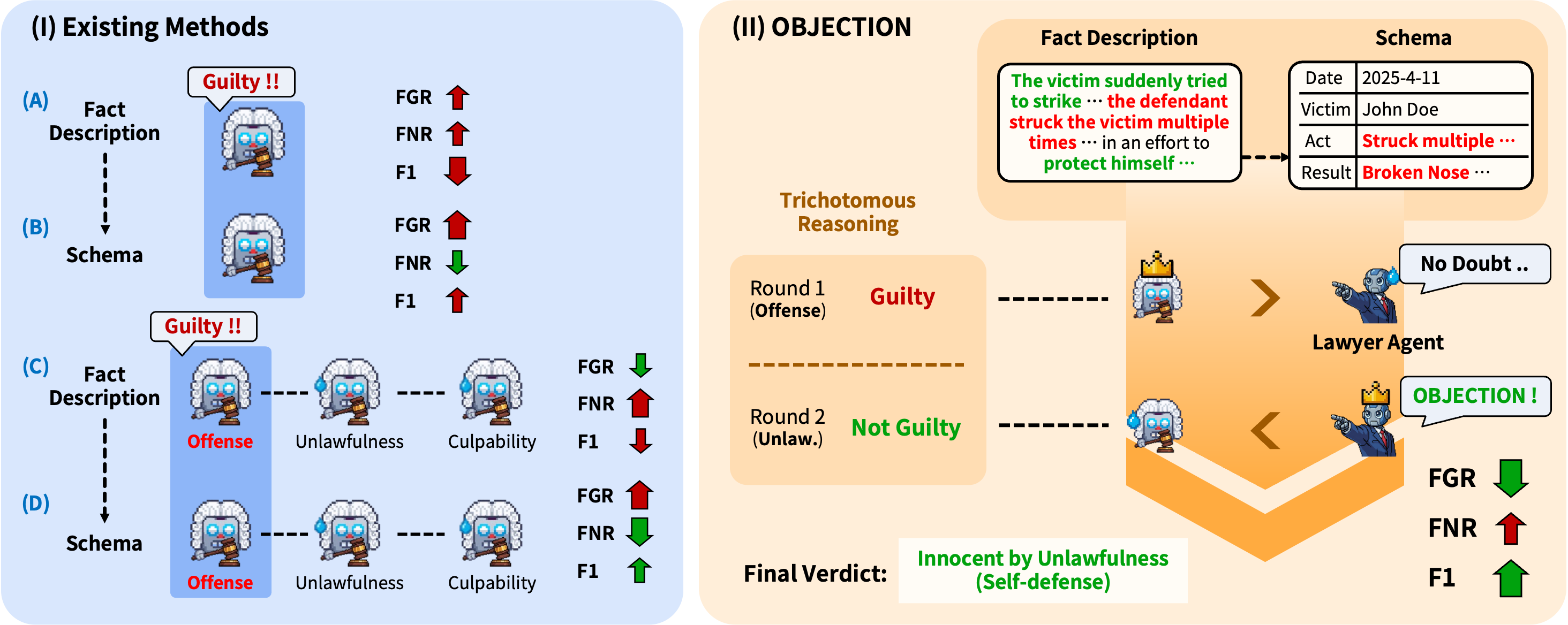}
  \caption {Comparative overview of LJP paradigms corresponding to Table~\ref{tab:prompt_results} and Sec.~\ref{sec:exp1}. FGR($\downarrow$) and FNR($\downarrow$) denote False Guilty Rate and False Not-Guilty Rate, respectively. {{Green arrow denotes improvements, while the reds denote degradations.}} Existing methods (A)–(D) suffer from guilty bias from fixed architectures. In contrast, (II) OBJECTION (Ours) employs a Lawyer Agent to actively dispute assumptions, efficiently mitigating bias while maintaining reasoning quality. The further reasoning round (e.g., Round 3) is skipped when a former legal doubt (e.g., Self-defense in Round 2) is established. For a detailed decision flow, see Fig.~\ref{fig:detail}.}
  \label{fig:overview}
\end{figure*}

We argue that guilty bias should be handled at inference time to neutralize bias effectively. We propose OBJECTION, a lawyer-agent-based critique-and-revise pipeline.
In contrast to generic critics \cite {madaan2023self}, the Lawyer Agent does not merely act as a consistency checker.
Instead, it operates under the principle of \emph{In Dubio Pro Reo}, actively excavating reasonable doubt hidden within the prosecutorial narrative to challenge overconfident guilt assumptions.
By integrating Schema-based Information Extraction (IE) and the 3-step reasoning structure into this agent, OBJECTION introduces a training-free pipeline that aligns with LLM inference while incorporating the adversarial nature of criminal law, effectively counteracting guilty bias. 

Our contributions are as follows:
\begin{itemize}[topsep=1pt, parsep=0.5pt]
    \item We quantify guilty bias in LJP systems using normative metrics such as False Guilty Rate (FGR), and show that prior LJP models exhibit a substantial inclination toward guilty predictions.
    \item We propose \textbf{OBJECTION}, an automatic judge framework with an adversarial lawyer agent without any extra training,
    and show that bias in existing frameworks can be neutralized by our proposed adversarial inference.
    \item We introduce a \textbf{Natural Innocent} dataset of real-world acquittal cases and reveal problems of existing approaches.
\end{itemize}

\section{Related Works}

\subsection{Legal Judgment Prediction and Guilty Bias}
Standard LJP settings utilize fact descriptions from written judgments, operating under the assumption that these often guilt-biased narratives represent neutral facts \cite{luo2017learning, zhong2018legal, xu2020ladan, yang2016hierarchical, xiao2021lawformer, chalkidis2020legalbert}. Many studies adopt Information Extraction (IE) or explicitly model Criminal Constitutive Elements to improve accuracy \cite{feng2022event, liu2025jurex}. These methods innovate in how they process the input but overlook the inherent bias within it, effectively inheriting the prosecutorial perspective.

Zhang et al. \cite{zhang2025ljpiv} were the first to explicitly identify this guilty bias and proposed synthetic innocent data and trichotomous training as remedies. While influential, this line of work handles bias primarily at training time, allowing models to rely on artificial cues of innocence. As a result, such models remain vulnerable to guilt-biased narratives at inference time, where no explicit signal of acquittal or adversarial challenge is present.

\begin{figure*}[ht]
  \includegraphics[width=1.0 \linewidth]{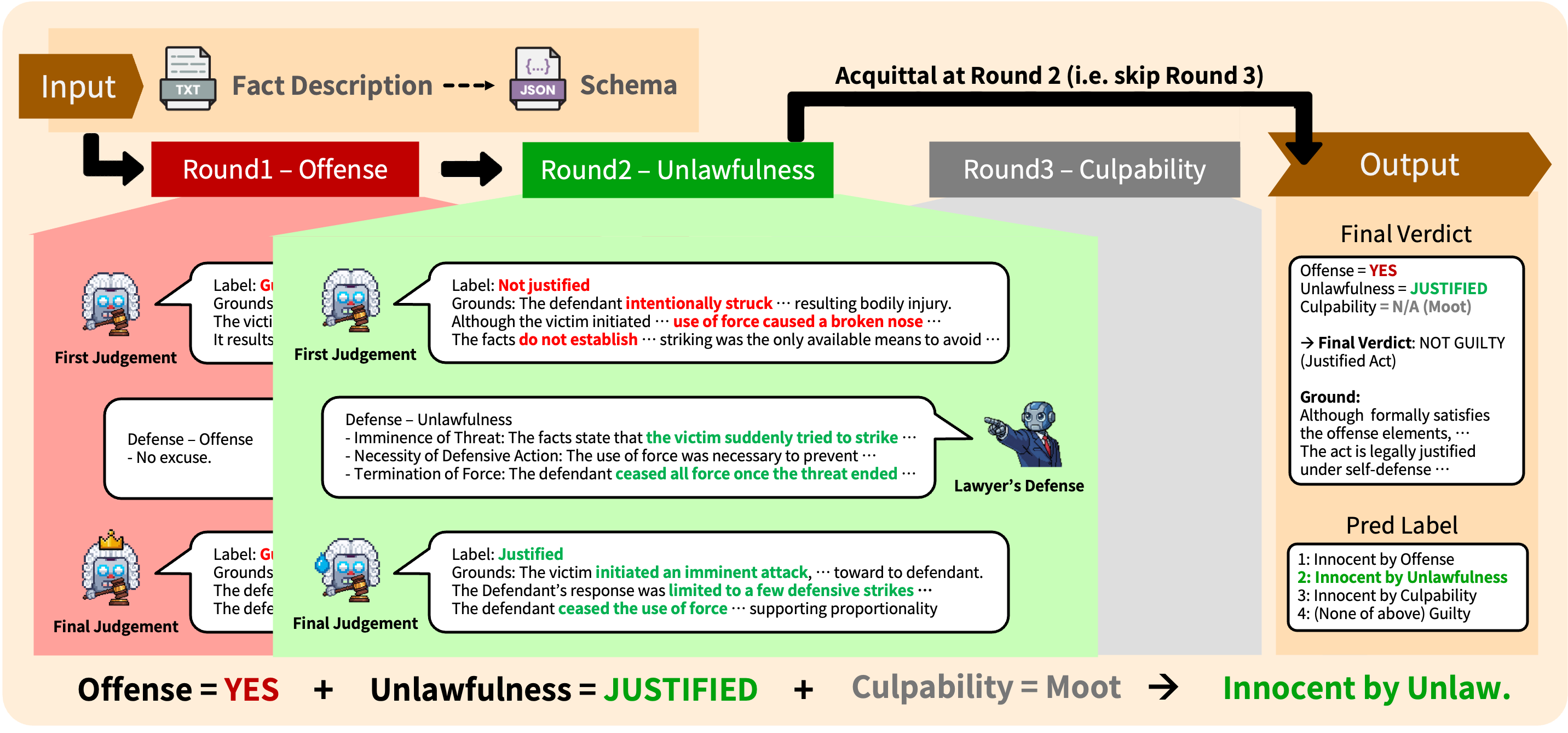}
  \caption {Detailed {{decision flow}} of the OBJECTION pipeline shown in Figure~\ref{fig:overview}-(II), illustrated with a self-defense case. The Lawyer Agent overturns the initial biased judgment in Round 2 by highlighting exculpatory details (e.g., imminence), triggering the early-exit mechanism for acquittal. This adversarial intervention effectively neutralizes narrative bias, reducing the False Guilty Rate (FGR) from 82.93\% to 16.69\% on the Natural Innocent dataset.}
  \label{fig:detail}
\end{figure*}

\subsection{Critique \& Revise Frameworks with Multi-agent Systems}
\label{multiagent_works}
Recent ``critique-and-revise'' frameworks utilize multi-agent pipelines, commonly favoring a ``divide-and-conquer'' strategy for granular feedback \cite{madaan2023self, chen2025magicore, xu2024llmrefine, zhou2024decrim, shinn2023reflexion}. However, these systems face a major constraint due to their iterative nature. Previous studies indicate such pipelines may amplify the model’s own biases \cite{xu2024pride}: rather than converging on accuracy, iterative refinement can reinforce the initial presumption of guilt.

This limitation is particularly critical in the legal domain. When the underlying LLM is already biased toward guilt, generic critique-and-revise or multi-agent debate systems \cite{chen2025debate} tend to reproduce and amplify this bias through iterative agreement, forming an echo chamber that reinforces the presumption of guilt. While recent approaches like AgentCourt \cite{chen-etal-2025-agentcourt} have introduced adversarial roles to simulate courtroom dynamics and improve legal question-answering, they focus primarily on civil domains and knowledge accumulation, leaving the fundamental issue of guilty bias in criminal judgments unaddressed.

In contrast, OBJECTION tackles this specific vulnerability by introducing a role-constrained adversarial Lawyer Agent whose sole objective is to construct the strongest possible defense. Unlike generic debaters or QA-oriented agents, our agent does not seek consensus or general knowledge refinement. Instead, it systematically enforces the presumption of innocence by actively injecting grounded, \textbf{reasonable doubts} at inference time. This effectively shatters the echo chamber, preventing bias amplification and enabling a genuine adversarial process that structurally corrects guilty bias; Appendix~\ref{abl:case_study} walks through a full case. Crucially, the Lawyer Agent is injected at each stage rather than after a completed verdict, and our Normal Critic ablation shows the effect stems from role asymmetry, not from adding a second agent (Table~\ref{tab:ablation_res}).

\section{Methodology}
The OBJECTION framework addresses the inherent guilty bias in Legal LLMs by orchestrating three core components: (1) Schema-based Information Extraction (IE) \cite{feng2022event, liu2025jurex}, (2) Trichotomous Reasoning with Issue Isolation \cite{zhang2025ljpiv}, and (3) Adversarial Argumentation. While existing strategies adopt IE or structural reasoning individually, they remain fragile against guilt-biased narratives. Specifically, LLMs tend to treat biased fact descriptions as neutral ground truths ($X_{fact}$), causing the reasoning process to collapse into a confirmation of guilt.

We propose a formal pipeline in which the Lawyer Agent serves as the orchestrator. Unlike a standard critic \cite{madaan2023self, shinn2023reflexion}, the Lawyer Agent serves as a binding mechanism that connects individually bias-fragile modules.

We instantiate a single LLM into two distinct roles via domain-specific prompting: the Judge ($\mathcal{M}$) and the Lawyer ($\mathcal{A}$). Instead of the standard prediction $\hat{y} = \mathcal{M}(X_{fact})$, we model the inference as a 3-step adversarial interaction process:
\vspace{-2.5pt}
\begin{equation}
    \hat{y} = \overrightarrow{\prod_{k \in T}} \Psi_{k} \left(\mathcal{M}, \mathcal{A}, X_{fact} \right)
\end{equation} 
where $T = \{\text{offense}, \text{unlawfulness}, \text{culpability}\}$ represents each step in the trichotomous reasoning structure, the arrow over $\prod$ denotes an ordered execution, and $\Psi_{k}$ denotes the sequential decision function refined by the adversarial intervention of $\mathcal{A}$ in step $k$. This design assures that the final verdict results from a legal debate rather than a unilateral confirmation of the prosecution's narrative.

\subsection{Input and Schema-based IE}
\label{SOAM}
Legal narratives are highly unstructured. While prior event extraction studies \cite{feng2022event, liu2025jurex} have made meaningful progress in structuring these narratives by decomposing them based on the Four Element Theory \cite{gao2019criminal}, applying these methods at scale often presents practical challenges. Because these approaches typically require defining specific rules for every charge, extending them across diverse crimes and legal systems requires substantial manual effort. Furthermore, while the four-element theory is fundamental in certain jurisdictions, other civil law countries (e.g., Germany, South Korea, Japan) heavily rely on the \textit{Tatbestand} concept, which explicitly separates objective and subjective elements. To overcome the dependency on charge-specific priors and provide a generalized factual abstraction, we propose a SOAM schema that captures the fundamental structure of criminal acts:
\begin{itemize}[topsep=5pt, parsep=0pt, itemsep=0.3pt]
    \item \textbf{Subject ($S$):} The actor or principal offender.
    \item \textbf{Object ($O$):} The target or victim of the act.
    \item \textbf{Actus Reus ($A$):} The strictly objective physical conduct itself and its consequences.
    \item \textbf{Mens Rea ($M$):} The subjective elements inferred from the context, rather than just explicit intent. We use the term broadly here.\footnote{We retain the term \textit{Mens Rea} to maintain alignment with the traditional \textit{Actus Reus} and \textit{Mens Rea} dichotomy. Here, it serves as a broad umbrella term encompassing various subjective states required for different offenses, consistent with broader criminal law jurisprudence.}
\end{itemize}

By abstracting facts into these four core components, our model performs extraction in a zero-shot manner (Prompt~\ref{prompt:soam}), effectively bypassing the need for granular annotation.

However, structure alone is not a panacea. As shown in our experiment (Table \ref{tab:prompt_results}), relying solely on the schema actually exacerbates the False Guilty Rate (FGR). Forcing the model to explicitly evaluate detailed elements without an adversarial counterbalance often reinforces its inherent prosecutorial bias. Both the schema and the 3-step reasoning are individually ``bias-fragile.''

Therefore, we utilize this extraction exclusively in the \textit{Offense} stage. While the schema alone may amplify bias, it provides a crucial \textit{structured factual anchor} for the Lawyer Agent. Without this S-O-A-M breakdown, the agent struggles to pinpoint specific legal weaknesses in unstructured narratives. By synergizing these components with an adversarial counterbalance, OBJECTION converts them into a robust bias-mitigation pipeline, while the subsequent \textit{Unlawfulness} and \textit{Culpability} stages access the raw description to capture subtle contextual nuances.

\subsection{Trichotomous Reasoning Framework}
We implement the dogmatic three-stage structure of criminal law prevalent in civil law systems \cite{dubber2005theories}.
The reasoning process follows a sequential chain $T = \{\text{offense}, \text{unlawfulness}, \text{culpability}\}$ adopting the logical flow proposed in recent LJP studies \cite{zhang2025ljpiv}:

\begin{enumerate}[label=\textbf{(\arabic*)}, topsep=5pt, parsep=0pt, itemsep=0.3pt]
    \item \textbf{Offense:} Does the act satisfy the constitutive elements of a crime?
    \item \textbf{Unlawfulness:} Are there justification grounds (e.g., self-defense) that negate unlawfulness?
    \item \textbf{Culpability:} Can the subject be held personally responsible? (e.g., insanity)
\end{enumerate}

The core of this pipeline is contextual isolation. Each step is separated to ensure logical integrity. For instance, when evaluating unlawfulness, the model must proceed on the basis that the offense is established, independent of culpability factors.

\subsection{Adversarial Lawyer Agent}
The main contribution of this work is the introduction of the Adversarial Lawyer Agent, which transcends the role of a standard self-critique agent \cite{madaan2023self, xu2024pride}.

\paragraph{Difference from Standard Critics}
While a standard critic passively verifies logical consistency (asking \textit{“Is the reasoning sound?”}), our Lawyer Agent actively shifts the burden of proof (similar to asking \textit{“Is there any room for reasonable doubt?”}). Even if the prosecution’s logic appears flawless on the surface, the Lawyer Agent is explicitly prompted to hypothesize exculpatory contexts (e.g., self-defense, intent negation), enforcing a rigorous adversarial review that generic critics cannot perform. In other words, while generic critics assess the validity of the model's logic, Lawyer Agent actively searches for a loophole that could be a reasonable doubt of guilt.

\paragraph{Adversarial Interaction Design}
Formally, we define the internal mechanism of $\Psi_k$ at each step as a dialectical interaction between $\mathcal{M}$ and $\mathcal{A}$:
\begin{align}
    j_k &= \mathcal{M}(X_{fact}), \quad d_k = \mathcal{A}(j_k, X_{fact}) \label{eq:intermediate_steps} \\
    \hat{y}_k &= \mathcal{M}(j_k, d_k, X_{fact}) \label{eq:final_verdict}
\end{align}

First, $\mathcal{M}$ produces an initial judgment $j_k$. Second, $\mathcal{A}$ inspects $j_k$ to generate a defense argument $d_k$ (Eq. \ref{eq:intermediate_steps}). Finally, this argument is injected back into the context, forcing $\mathcal{M}$ to re-evaluate the case to derive the final stage verdict $\hat{y}_k$ (Eq. \ref{eq:final_verdict}).

This formulation highlights that the final decision is explicitly conditioned on the defense narrative, structurally blocking the model's inherent guilty bias. Our design of interaction systematically implements the \textit{Exercise of Defense Rights} in criminal procedure.

\subsection{Sequential Verdict Determination}
The final verdict is determined by synthesizing the results of three stages. In accordance with the hierarchy of criminal law, the pipeline employs an Early Exit Mechanism: if a higher-level stage is negated (e.g., Offense is not established), the process terminates immediately with an acquittal, without evaluating subsequent stages. This ensures both computational efficiency and logical consistency by preventing error propagation.

Unlike simple binary classification, our system produces Granular Acquittal Labels (e.g., \textit{Innocent by Offense}, \textit{Unlawfulness}, \textit{Culpability}). This granularity significantly improves the explainability of the judgment, clarifying the precise legal grounds for the verdict.

\section{Experiments}
\subsection{Experimental Setup}
\paragraph{Datasets}
We evaluate our framework on three different datasets to verify both in-domain performance and cross-domain generalizability.
\begin{itemize}[topsep=5pt, parsep=0pt, itemsep=0.3pt]
    \item \textbf{LJPIV-CAIL:} A benchmark dataset \cite{zhang2025ljpiv} containing criminal cases with synthetic innocent samples generated via counterfactual rewriting from the CAIL dataset \cite{xiao2018cail}.
    \item \textbf{LJPIV-ELAM:} An out-of-domain dataset \cite{zhang2025ljpiv} based on ELAM \cite{yu2022explainable}, constructed via the same generation pipeline to evaluate cross-domain generalization.
    \item \textbf{Natural Innocent (Ours):} To resolve the limitations of synthetic innocence, we introduce a comprehensive dataset consisting of \textbf{3,412 real criminal judgments} from South Korean criminal courts. Detailed information can be found in Appendix~\ref{sec:dataset_detail}.
\end{itemize}

\paragraph{Baselines} 
We categorize our baselines into two groups to identify the individual effects of prompting strategies and architectural choices. All inference experiments utilize \texttt{Qwen-2.5-7B-Instruct} \cite{qwen2.5}, \texttt{Llama-3.1-8B-Instruct} \cite{dubey2024llama}, and \texttt{Gemma-2-9B-it} \cite{gemma_2024} as the backbones.

\paragraph{Prompting Baselines}
To investigate whether simply adding Schema or Structure mitigates bias, we designed four incremental prompting strategies.
\begin{enumerate}[label=\textbf{(\Alph*)}, topsep=5pt, parsep=0pt, itemsep=0.3pt]
    \item \textbf{Plain LLM: } Performs direct zero-shot prediction using standard prompts. 
    \item \textbf{Schema-based: } Utilizes the extracted SOAM elements to focus on factual components.
    \item \textbf{Naive Trichotomy: } Compels the model to follow the 3-step reasoning structure. 
    \item \textbf{Schema + Trichotomy: } Naively combines both the schema and the 3-step structure.
\end{enumerate}

\paragraph{Main Baselines}
We compare OBJECTION against \textbf{LJPIV} \cite{zhang2025ljpiv} 
(fine-tuned SOTA, in-domain upper bound), \textbf{Self-Refine} 
\cite{madaan2023self} (generic critic without legal persona), and 
\textbf{Debate-Feedback} \cite{chen2025debate} (round-table debate 
via stochastic sampling).

\paragraph{Metrics}
We report \textbf{G-F1} and \textbf{NG-F1}, which represent the F1 scores for the Guilty and Not-Guilty classes, respectively. We also measure the macro Precision/Recall/F1 of the 3-step reasoning process. Crucially, beyond standard performance metrics, we introduce normative error metrics that directly reflect legal risk:
\begin{itemize}[topsep=5pt, parsep=0pt, itemsep=0.3pt]
    \item \textbf{False Guilty Rate (FGR):} The ratio of innocent cases predicted as guilty ($FP / (TN + FP)$). A high FGR indicates a violation of the presumption of innocence.
    \item \textbf{False Not-Guilty Rate (FNR):} The ratio of guilty cases predicted as innocent ($FN / (TP + FN)$).
\end{itemize}

\begin{table}[h]
\centering
\footnotesize
\caption{Impact of adversarial intervention. (A)--(D) correspond to the existing methods in Fig.~\ref{fig:overview}. The Naive combination \textbf{(D)} exhibits severe guilty bias despite high accuracy. Integrating the \textbf{Lawyer Agent} reduces FGR (53.21\% $\rightarrow$ 10.71\%) while improving reasoning.}
\label{tab:prompt_results}
\resizebox{\columnwidth}{!}{%
\begin{tabular}{lccccc}
\toprule
\textbf{Method} & \textbf{G-F1} & {\textbf{NG-F1}} & \textbf{Macro F1} & \textbf{FGR}$\downarrow$ & \textbf{FNR}$\downarrow$ \\
\midrule
(A) Plain LLM & 0.37 & 0.58 & 0.48 & 29.64 & 70.36 \\
(B) Schema & 0.67 & 0.53 & 0.60 & 56.07 & 21.79 \\
(C) Naive Tri. & 0.31 & 0.63 & 0.47 & 17.86 & 78.21 \\
(D) Schema+Tri. & \textbf{0.77} & 0.62 & 0.70 & 53.21 & \textbf{3.93} \\
\midrule
(D) + Lawyer & 0.75 & \textbf{0.80} & \textbf{0.78} & \textbf{10.71} & 33.93 \\
\bottomrule
\end{tabular}%
}
\end{table}

\subsection{Experiment 1: Limitations of Naive Prompting} \label{sec:exp1}

\begin{table*}[ht!]
\centering
\footnotesize
\renewcommand{\arraystretch}{0.7}      
\caption{Main Comparative Results. \textbf{Bold} indicates the best performance within each group. The results show that OBJECTION achieves a consistent directional effect across diverse backbones. While \textbf{Debate-Feedback} exhibits severe volatility depending on the backbone architecture (e.g., uncontrolled over-acquittal on Llama), OBJECTION consistently mitigates guilty bias (low FGR) while preserving a solid conviction capability (Macro-F1). $^{*}$ denotes datasets whose innocent cases are synthetic, while $^{\dagger}$ denotes datasets whose innocent cases are real-world acquittals.}

\label{tab:main_results_unified}
\resizebox{\textwidth}{!}{%
\begin{tabular}{l l ccc ccc cc}
\toprule
\multirow{2}{*}{\textbf{Dataset}} & \multirow{2}{*}{\textbf{Backbone \& Method}} & \multirow{2}{*}{\textbf{G-F1}} & \multirow{2}{*}{{\textbf{NG-F1}}} & \multirow{2}{*}{\textbf{Macro-F1}} & \multicolumn{3}{c}{\textbf{3-Step Reasoning}} & \multirow{2}{*}{\textbf{FGR (\%)}$\downarrow$} & \multirow{2}{*}{\textbf{FNR (\%)}$\downarrow$} \\
\cmidrule(lr){6-8}
 & & & & & \textbf{Prec.} & \textbf{Recall} & \textbf{F1} & & \\
\midrule

\multirow{19}{*}{\shortstack{\textbf{LJPIV-}\\\textbf{CAIL}$^{*}$}} 
 & \quad LJPIV (Fine-tuned) & 0.80 & 0.83 & 0.82 & 0.62 & 0.71 & 0.61 & 9.29 & 27.14 \\
 \cmidrule(lr){2-10}
 & \textbf{Qwen-2.5-7B} & & & & & & & & \\
 & \quad Baseline D & \textbf{0.77} & 0.62 & 0.70 & 0.54 & 0.40 & 0.39 & 53.21 & \textbf{3.93} \\
 & \quad Self-Refine & \textbf{0.77} & 0.64 & 0.71 & 0.66 & 0.56 & 0.57 & 49.29 & 6.79 \\
 & \quad Debate-Feedback & 0.74 & 0.76 & 0.75 & 0.70 & 0.61 & 0.62 & 22.14 & 27.50 \\
 & \quad \textbf{OBJECTION (Ours)} & 0.75 & \textbf{0.80} & \textbf{0.78} & \textbf{0.87} & \textbf{0.81} & \textbf{0.82} & \textbf{10.71} & 33.93 \\
 \cmidrule(lr){2-10}
 & \textbf{Llama-3.1-8B} & & & & & & & & \\
 & \quad Baseline D & \textbf{0.76} & 0.56 & 0.66 & \textbf{0.83} & 0.57 & 0.58 & 60.36 & \textbf{2.50} \\
 & \quad Self-Refine & 0.69 & 0.44 & 0.57 & 0.74 & 0.59 & 0.60 & 68.93 & 10.36 \\
 & \quad Debate-Feedback & 0.58 & \textbf{0.72} & 0.65 & 0.60 & 0.61 & 0.60 & \textbf{20.36} & 52.86 \\
 & \quad \textbf{OBJECTION (Ours)} & 0.73 & 0.67 & \textbf{0.70} & 0.73 & \textbf{0.69} & \textbf{0.69} & 40.36 & 18.57 \\
 \cmidrule(lr){2-10}
 & \textbf{Gemma-2-9B} & & & & & & & & \\
 & \quad Baseline D & 0.77 & 0.59 & 0.68 & 0.76 & 0.58 & 0.60 & 57.50 & 1.07 \\
 & \quad Self-Refine & 0.77 & 0.57 & 0.67 & \textbf{0.87} & 0.62 & \textbf{0.65} & 60.00 & \textbf{0.36} \\
 & \quad Debate-Feedback & 0.81 & 0.80 & 0.81 & 0.71 & 0.62 & 0.63 & \textbf{22.50} & 16.43 \\
 & \quad \textbf{OBJECTION (Ours)} & \textbf{0.84} & \textbf{0.80} & \textbf{0.82} & 0.67 & \textbf{0.64} & \textbf{0.65} & 26.43 & 9.29 \\
\midrule

\multirow{19}{*}{\shortstack{\textbf{LJPIV-}\\\textbf{ELAM}$^{*}$}} 
 & \quad LJPIV (Fine-tuned) & 0.80 & 0.68 & 0.74 & 0.87 & 0.66 & 0.71 & 47.60 & 2.40 \\
 \cmidrule(lr){2-10}
 & \textbf{Qwen-2.5-7B} & & & & & & & & \\
 & \quad Baseline D & 0.75 & 0.58 & 0.67 & 0.51 & 0.38 & 0.37 & 56.40 & 5.60 \\
 & \quad Self-Refine & 0.76 & 0.58 & 0.67 & 0.75 & 0.55 & 0.59 & 57.60 & \textbf{3.20} \\
 & \quad Debate-Feedback & \textbf{0.80} & 0.74 & 0.77 & \textbf{0.87} & 0.65 & 0.67 & 27.00 & 15.00 \\
 & \quad \textbf{OBJECTION (Ours)} & 0.78 & \textbf{0.77} & \textbf{0.78} & 0.82 & \textbf{0.80} & \textbf{0.81} & \textbf{23.20} & 21.60 \\
 \cmidrule(lr){2-10}
 & \textbf{Llama-3.1-8B} & & & & & & & & \\
 & \quad Baseline D & \textbf{0.76} & 0.53 & 0.65 & 0.52 & 0.54 & 0.52 & 64.00 & \textbf{0.40} \\
 & \quad Self-Refine & 0.66 & 0.33 & 0.50 & 0.67 & 0.55 & 0.54 & 77.60 & 12.00 \\
 & \quad Debate-Feedback & 0.61 & \textbf{0.74} & \textbf{0.68} & 0.56 & 0.54 & 0.54 & \textbf{23.20} & 56.00 \\
 & \quad \textbf{OBJECTION (Ours)} & 0.71 & 0.64 & \textbf{0.68} & \textbf{0.73} & \textbf{0.69} & \textbf{0.67} & 41.60 & 22.00 \\
 \cmidrule(lr){2-10}
 & \textbf{Gemma-2-9B} & & & & & & & & \\
 & \quad Baseline D & 0.75 & 0.51 & 0.63 & 0.81 & 0.56 & 0.58 & 66.00 & \textbf{0.00} \\
 & \quad Self-Refine & 0.73 & 0.41 & 0.57 & \textbf{0.84} & 0.54 & 0.56 & 74.40 & \textbf{0.00} \\
 & \quad Debate-Feedback & 0.80 & 0.72 & 0.76 & 0.73 & 0.63 & 0.67 & 38.40 & 9.60 \\
 & \quad \textbf{OBJECTION (Ours)} & \textbf{0.83} & \textbf{0.77} & \textbf{0.80} & 0.67 & \textbf{0.68} & \textbf{0.68} & \textbf{36.60} & 2.00 \\

\midrule

\multirow{19}{*}{\shortstack{\textbf{Natural}\\\textbf{Innocent}$^{\dagger}$}} 
 & \quad LJPIV (Fine-tuned) & 0.35 & 0.29 & 0.32 & 0.47 & 0.06 & 0.11 & 82.93 & 50.12 \\
 \cmidrule(lr){2-10}
 & \textbf{Qwen-2.5-7B} & & & & & & & & \\
 & \quad Baseline D & 0.55 & 0.27 & 0.41 & 0.48 & 0.34 & 0.21 & 84.49 & 9.99 \\
 & \quad Self-Refine & 0.57 & 0.23 & 0.40 & 0.54 & 0.36 & 0.24 & 86.28 & \textbf{2.87} \\
 & \quad Debate-Feedback & \textbf{0.71} & 0.73 & 0.72 & 0.47 & 0.48 & 0.44 & 42.95 & 5.34 \\
 & \quad \textbf{OBJECTION (Ours)} & 0.70 & \textbf{0.82} & \textbf{0.76} & \textbf{0.57} & \textbf{0.48} & \textbf{0.47} & \textbf{16.69} & 31.29 \\
 \cmidrule(lr){2-10}
 & \textbf{Llama-3.1-8B} & & & & & & & & \\
 & \quad Baseline D & 0.51 & 0.66 & 0.58 & 0.46 & 0.42 & 0.38 & 36.73 & 45.08 \\
 & \quad Self-Refine & 0.53 & 0.33 & 0.43 & 0.43 & 0.41 & 0.34 & 78.55 & \textbf{16.89} \\
 & \quad Debate-Feedback & 0.41 & \textbf{0.78} & 0.59 & 0.46 & 0.44 & 0.41 & \textbf{7.69} & 70.72 \\
 & \quad \textbf{OBJECTION (Ours)} & \textbf{0.63} & 0.65 & \textbf{0.64} & \textbf{0.51} & \textbf{0.50} & \textbf{0.49} & 46.39 & 19.05 \\
 \cmidrule(lr){2-10}
 & \textbf{Gemma-2-9B} & & & & & & & & \\
 & \quad Baseline D & 0.57 & 0.18 & 0.38 & 0.53 & 0.41 & 0.32 & 90.19 & 1.16 \\
 & \quad Self-Refine & \textbf{0.60} & 0.34 & 0.47 & \textbf{0.62} & 0.46 & 0.39 & 79.73 & \textbf{0.15} \\
 & \quad Debate-Feedback & 0.49 & 0.69 & 0.59 & 0.50 & 0.48 & 0.47 & 37.01 & 50.74 \\
 & \quad \textbf{OBJECTION (Ours)} & \textbf{0.60} & \textbf{0.73} & \textbf{0.67} & 0.52 & \textbf{0.51} & \textbf{0.50} & \textbf{29.89} & 36.10 \\
\bottomrule
\end{tabular}%
}
\end{table*}

Table \ref{tab:prompt_results} highlights the limitations of naive prompting strategies and the necessity of adversarial intervention. Approaches such as Schema-based extraction (B) and Naive Trichotomy (C) do not achieve a balance between accuracy and fairness. Even when combined (D), these methods produce logically structured yet significantly biased predictions, as indicated by an FGR of 53.21\%.

In contrast, incorporating the Lawyer Agent into this pipeline ((D)+Lawyer) reduces the FGR to 10.71\%. These results indicate that a structured reasoning framework alone is insufficient and active adversarial reasoning is essential to mitigate bias.

\subsection{Experiment 2: Main Comparative Results}

Table \ref{tab:main_results_unified} compares OBJECTION with main baselines. We examine the results, focusing on mitigating guilty bias and cross-domain performance.

\paragraph{Baselines vs. Generalization} Existing methods encounter problems with stability and generalization across different LLM architectures. LJPIV attains high in-domain accuracy but suffers from severe guilty bias on out-of-domain datasets such as ELAM (FGR 47.60\%). Crucially, Debate-Feedback shows substantial volatility depending on the backbone model. For instance, it fails to effectively mitigate bias on Qwen-2.5 (e.g., CAIL FGR 22.14\% vs. 10.71\%), experiences uncontrolled over-acquittal on Llama-3.1 (e.g., 56.00\% FNR on ELAM), and exhibits inconsistent bias control on Gemma-2. These fluctuations indicate that debate through stochastic sampling lacks a reliable steering mechanism. By comparison, OBJECTION demonstrates a consistent directional effect, reducing guilty bias on most backbones and achieving the highest overall accuracy (Macro-F1) across backbones; significance tests are reported in Appendix~\ref{sec:significance}.

\paragraph{Structural Integrity and Reasoning} The superior generalization of OBJECTION is attributed to its rigorous adherence to legal procedure. As indicated in the ``3-Step Reasoning'' columns, this method maintains high element-wise detection scores (F1: 0.82 on CAIL, 0.81 on ELAM) across domains. In contrast, Baseline D fails on the ELAM dataset (F1 0.37), demonstrating that, without the Lawyer Agent's guidance, the model loses logical coherence.

\paragraph{Synthetic vs. Natural Innocence} The most distinct performance disparity is observed in the Natural Innocent dataset. LJPIV performs poorly (FGR 82.93\%) as a result of overfitting to synthetic innocent cues. Debate-Feedback also underperforms in this context; when using the Qwen backbone, it produces a high FGR of 42.95\%, indicating a failure to protect innocent defendants. In contrast, OBJECTION exhibits greater robustness, achieving an FGR of 16.69\% (Qwen). These results indicate that the Lawyer Agent effectively identifies reasonable doubt concealed within the narrative, enabling the detection of innocence without dependence on explicit shortcuts; a breakdown by acquittal ground is given in Appendix~\ref{sec:coverage}.

\begin{table}[h]
\centering
\small
\caption{Ablation Results (w/ \texttt{Qwen 2.5}) {{on ELAM dataset.}} The Lawyer Agent is effective in a low False Guilty Rate (FGR). {Similar ablation trends are observed on the CAIL dataset where LJPIV was originally trained; see Appendix~\ref{sec:cail_ablation} for details.}}
\label{tab:ablation_res}
\resizebox{\columnwidth}{!}{%
\begin{tabular}{lcccc}
\toprule
\textbf{Configuration} & \textbf{G-F1} & {\textbf{NG-F1}} & \textbf{FGR}$\downarrow$ & \textbf{FNR}$\downarrow$ \\
\midrule
\textbf{OBJECTION (Full)} & 0.78 & \textbf{0.77} & 23.20 & 21.60 \\
(-) Lawyer Agent & \textbf{0.83} & 0.75 & 38.80 & \textbf{2.00} \\
(-) Schema & 0.52 & 0.72 & \textbf{8.00} & 62.40 \\
(-) 3-Step Structure & 0.38 & 0.56 & 34.80 & 68.00 \\
Ours w/ Normal Critic & 0.77 & 0.60 & 56.40 & \textbf{2.00} \\
\midrule
LJPIV & 0.80 & 0.68 & 47.60 & 2.40 \\
LJPIV + Lawyer Agent & 0.01 & 0.67 & 0.40 & 99.60 \\
\bottomrule
\end{tabular}%
}
\end{table}

\subsection{Ablation Studies and Analysis} \label{sec:analysis}
\paragraph{Ablation Study}

To validate the contribution of each component, we conducted an ablation study (Table \ref{tab:ablation_res}).

\begin{itemize}[topsep=5pt, parsep=0pt, itemsep=0.3pt]
    \item \textbf{Role of Lawyer Agent:} Removing the Lawyer Agent causes FGR to spike from 23.20\% to 38.80\%, while NG-F1 drops from 0.77 to 0.75. Notably, G-F1 rises to 0.83 in this setting, above the full pipeline. Given the guilty bias established in Section~\ref{sec:exp1}, the Lawyer Agent acts as a counterweight rather than a reasoning enhancer: it shifts where the decision boundary sits, making verdicts fairer rather than sharper.
    \item \textbf{Role of Structural Anchors:} Removing either the SOAM schema (factual anchor) or the 3-step structure (logical anchor) results in high FNRs (62.40\% and 68.00\%, respectively). Without these constraints to ground the debate, the Lawyer Agent becomes overly dominant and blindly argues for innocence, leading to excessive acquittals.
    \item \textbf{Ours w/ Normal Critic:} Replacing the Lawyer Agent with a generic critic yields a high guilty bias (FGR 56.40\%). Without a specific mandate to establish reasonable doubt, LLMs act as compliant approvers of the input narrative rather than adversarial challengers.
    \item \textbf{Fine-tuning + Lawyer:} Adding our Lawyer Agent to the fine-tuned LJPIV leads to failure (FNR 99.60\%). Since LJPIV has already learned to detect synthetic innocence, the additional adversarial intervention amplifies doubt excessively, creating an ``echo chamber'' \cite{xu2024pride} that leads to indiscriminate acquittals.
\end{itemize}

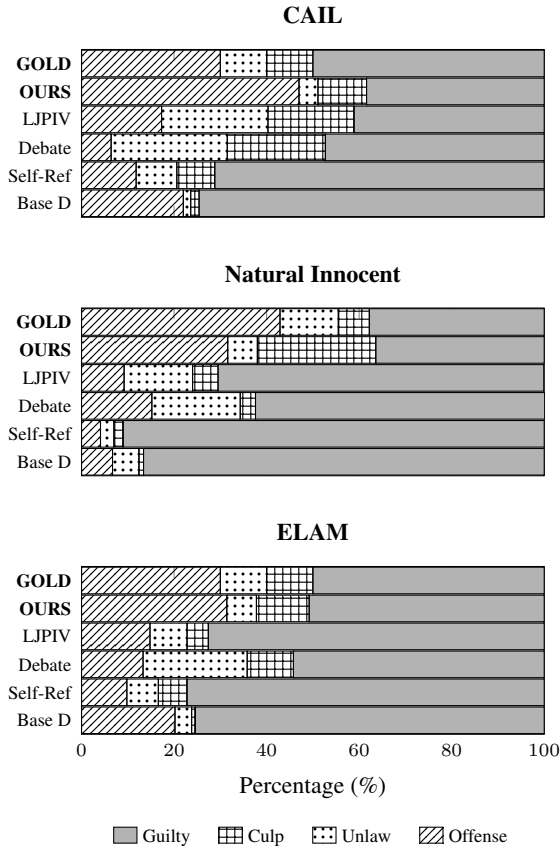
\begin{figure}[h!]
    \centering
    \begin{tikzpicture}
        \begin{groupplot}[
            group style={
                group size=1 by 3,
                vertical sep=1.2cm,
                x descriptions at=edge bottom,
            },
            xbar stacked,
            width=\linewidth,
            height=3.8cm, 
            ytick={1, 2, 3, 4, 5, 6},
            yticklabels={Base D, Self-Ref, Debate, LJPIV, \textbf{OURS}, \textbf{GOLD}},
            xmin=0, xmax=100,
            xlabel={Percentage (\%)},
            ytick style={draw=none},
            tick label style={font=\scriptsize},
            label style={font=\footnotesize},
            legend style={
                at={(0.5,-0.5)},
                anchor=north, 
                legend columns=-1, 
                draw=none, 
                font=\scriptsize,
                /tikz/every even column/.append style={column sep=0.3cm}
            },
            reverse legend,
        ]
        \nextgroupplot[title={\small \textbf{CAIL}}]
            \addplot[pattern=north east lines] coordinates {
                (22.0,1) (11.8,2) (6.4,3) (17.3,4) (47.0,5) (30.0,6)
            };
            \addplot[pattern=dots] coordinates {
                (1.6,1) (8.8,2) (25.1,3) (23.0,4) (4.1,5) (10.0,6)
            };
            \addplot[pattern=grid] coordinates {
                (1.8,1) (8.2,2) (21.2,3) (18.6,4) (10.5,5) (10.0,6)
            };
            \addplot[fill=black!30] coordinates {
                (74.6,1) (71.2,2) (47.3,3) (41.1,4) (38.4,5) (50.0,6)
            };

        \nextgroupplot[title={\small \textbf{Natural Innocent}}]
            \addplot[pattern=north east lines] coordinates {
                (6.7,1) (4.1,2) (15.2,3) (9.2,4) (31.6,5) (42.9,6)
            };
            \addplot[pattern=dots] coordinates {
                (5.7,1) (2.9,2) (19.1,3) (14.8,4) (6.4,5) (12.6,6)
            };
            \addplot[pattern=grid] coordinates {
                (1.0,1) (2.0,2) (3.3,3) (5.5,4) (25.6,5) (6.7,6)
            };
            \addplot[fill=black!30] coordinates {
                (86.6,1) (91.0,2) (62.5,3) (70.4,4) (36.4,5) (37.8,6)
            };

        \nextgroupplot[title={\small \textbf{ELAM}}]
            \addplot[pattern=north east lines] coordinates {
                (20.2,1) (9.8,2) (13.3,3) (14.8,4) (31.4,5) (30.0,6)
            };
            \addplot[pattern=dots] coordinates {
                (3.6,1) (6.8,2) (22.5,3) (8.0,4) (6.4,5) (10.0,6)
            };
            \addplot[pattern=grid] coordinates {
                (0.8,1) (6.2,2) (10.0,3) (4.6,4) (11.4,5) (10.0,6)
            };
            \addplot[fill=black!30] coordinates {
                (75.4,1) (77.2,2) (54.2,3) (72.6,4) (50.8,5) (50.0,6)
            };

            \legend{Offense, Unlaw, Culp, Guilty}
        \end{groupplot}
    \end{tikzpicture}
    \caption{Prediction distributions across datasets. Baselines over-predict the ``Guilty'' class (gray bars), whereas OBJECTION closely matches the GOLD standard by accurately identifying specific grounds for acquittal.}
    \label{fig:all_dist_gold}
\end{figure}

\paragraph{Distribution Analysis}

We analyzed the prediction distribution (Figure \ref{fig:all_dist_gold}). Baselines like Baseline D and Self-Refine skew heavily towards the ``Guilty'' class (Gray bars), diverging significantly from the GOLD standard. In contrast, OBJECTION aligns most closely with the GOLD distribution across all datasets. This indicates that our model does not merely ``guess'' innocence to lower FGR. Instead, it accurately maintains necessary guilty convictions while correctly classifying the specific legal grounds for acquittal (Offense vs. Unlawfulness vs. Culpability), demonstrating a highly calibrated understanding of legal liability.

\begin{table}[h]
\centering
\small
\setlength{\tabcolsep}{4pt}
\caption{Robustness on Commercial APIs compared to baselines. While Debate-Feedback offers moderate bias reduction, OBJECTION consistently attains fairness (low FGR) and reasoning quality (high 3-Step F1) across both GPT-4.1 and Gemini architectures.}
\label{tab:commercial_analysis}
\resizebox{\columnwidth}{!}{%
\begin{tabular}{l l cc cc}
\toprule
\multirow{2}{*}{\textbf{Model}} & \multirow{2}{*}{\textbf{Method}} & \multicolumn{2}{c}{\textbf{CAIL}} & \multicolumn{2}{c}{\textbf{Natural}} \\
\cmidrule(lr){3-4} \cmidrule(lr){5-6}
 & & \textbf{FGR}$\downarrow$ & \textbf{3-S F1}$\uparrow$ & \textbf{FGR}$\downarrow$ & {\textbf{3-S F1}$\uparrow$} \\
\midrule
\multirow{3}{*}{\shortstack{\textbf{GPT-4.1}\\\textbf{mini}}} 
 & Baseline D & 40.36 & 0.54 & 59.71 & 0.08 \\
 & Debate. & 30.84 & 0.66 & \textbf{54.92} & 0.22 \\
 & \textbf{OBJECTION} & \textbf{21.43} & \textbf{0.75} & 58.46 & \textbf{0.34} \\
\midrule
\multirow{3}{*}{\shortstack{\textbf{Gemini}\\\textbf{2.5-Flash}}} 
 & Baseline D & 51.07 & 0.56 & 76.75 & 0.03 \\
 & Debate. & 42.63 & 0.63 & 68.41 & 0.17 \\
 & \textbf{OBJECTION} & \textbf{36.07} & \textbf{0.65} & \textbf{61.95} & \textbf{0.33} \\
\bottomrule
\end{tabular}%
}
\end{table}

\paragraph{Generalization to Commercial APIs}

To assess scalability, OBJECTION was applied to commercial APIs (\texttt{GPT-4.1-mini}, \texttt{Gemini-2.5-Flash}). As presented in Table \ref{tab:commercial_analysis}, the framework consistently reduces guilty bias on the CAIL dataset, lowering the FGR to 21.43\%. Furthermore, on the Natural Innocent benchmark, OBJECTION achieves a substantial improvement in 3-Step Reasoning F1 (0.34), significantly surpassing the baseline (0.08). These findings show that the pipeline effectively integrates the three-stage judgment logic of criminal law into general-purpose large language models, ensuring that acquittals are grounded in valid legal reasoning.

\begin{table}[h]
\centering
\small
\setlength{\tabcolsep}{3.5pt}
\caption{Controllability via Persona Design. Unlike uncontrolled baselines, OBJECTION allows explicit steering through distinct lawyer variants. Our experiments reveal a trade-off between fairness and accuracy, demonstrating that OBJECTION can actively steer this trade-off by modulating defense intensity.}
\label{tab:soft_variant}
\resizebox{\columnwidth}{!}{%
\begin{tabular}{l cc cc cc}
\toprule
\multirow{2}{*}{\textbf{Method}} & \multicolumn{2}{c}{\textbf{CAIL}} & \multicolumn{2}{c}{\textbf{ELAM}} & \multicolumn{2}{c}{\textbf{Natural}} \\
\cmidrule(lr){2-3} \cmidrule(lr){4-5} \cmidrule(lr){6-7}
 & \textbf{G-F1} & \textbf{FGR} & \textbf{G-F1} & \textbf{FGR} & {\textbf{NG-F1}} & \textbf{FGR} \\
\midrule
LJPIV & 0.80 & 9.29 & 0.80 & 47.60 & 0.29 & 82.93 \\
\midrule
Baseline D & 0.77 & 53.21 & 0.75 & 56.40 & 0.27 & 84.49 \\
Debate. & 0.74 & 22.14 & 0.80 & 27.00 & 0.73 & 42.95 \\
\textbf{OBJECTION} & 0.75 & \textbf{10.71} & 0.78 & \textbf{23.20} & \textbf{0.82} & \textbf{16.69} \\
\textbf{OBJECTION-Soft} & \textbf{0.83} & 31.07 & \textbf{0.84} & 31.60 & 0.58 & 58.84 \\
\bottomrule
\end{tabular}%
}
\end{table}

\subsection{Controllability: The ``Soft'' Variant}
\label{sec:soft_variant}
A key advantage of our framework is the ability to explicitly control the model's behavior by modulating the Lawyer Agent's intensity. This stands in sharp contrast to debate-based methods \cite{chen2025debate}, where response diversity relies on stochastic sampling with identical prompts, lacking a mechanism to systematically control the ``degree of doubt''. By designing a softened persona (OBJECTION-Soft), we provide a deterministic ``steering knob'' that allows users to control the trade-off between conviction accuracy and the presumption of innocence. 

Table \ref{tab:soft_variant} provides empirical validation of this steerability. OBJECTION-Soft attains higher conviction accuracy (G-F1 0.83 on CAIL) compared to the standard version (0.75), but this improvement is accompanied by an increased False Guilty Rate (31.07\%). Additionally, the Soft variant's decline in performance on the Natural Innocent benchmark (FGR 58.84\%) indicates that real-world acquittals involving subtle implicit cues require stronger adversarial intervention, emphasizing the need for a tunable framework for domain-adaptive deployment. This FGR--FNR asymmetry is deliberate. A false conviction is irreversible whereas a false acquittal is correctable (\textit{in dubio pro reo}), and across our settings the reduction in FGR is substantially larger than the accompanying rise in FNR.

\subsection{Human Expert Evaluation}
\label{sec:human_eval}

To rigorously assess the practical applicability and legal validity of the generated defense arguments, we conducted a qualitative evaluation with human experts. The evaluation panel consisted of active law enforcement investigators with extensive experience in criminal investigations.

\paragraph{Evaluation Setup} 
We randomly sampled 50 cases from the Natural Innocent dataset. The experts blindly evaluated the defense arguments generated by our Lawyer Agent and the debater agents in the Debate-Feedback pipeline. Each argument was scored on a 5-point Likert scale (1: Poor, 5: Excellent) across four criteria:
\begin{itemize}[topsep=2pt, parsep=0pt, itemsep=0pt, leftmargin=*]
    \item \textbf{Legal Relevance (Rel.):} Legal validity and appropriateness.
    \item \textbf{Argument Strength (Str.):} Persuasiveness and logical coherence.
    \item \textbf{Factual Grounding (Grd.):} Strict anchoring to the provided input facts.
    \item \textbf{Human Similarity (Sim.):} Resemblance to a professional legal opinion.
\end{itemize}

\begin{table}[h!]
    \centering
    \small
    \caption{\textbf{Human Expert Evaluation Results ($N=50$).} (A) Our Lawyer Agent outperforms the baseline across all metrics. (B) Step-wise analysis reveals that the agent performs best in the Unlawfulness stage (Avg 4.57), verifying its strong logical reasoning capability.}
    \label{tab:human_eval_compare}
    \resizebox{\columnwidth}{!}{%
    \begin{tabular}{lccccc}
        \toprule
        \multicolumn{6}{c}{\textbf{(A) Comparative Analysis (Baseline vs. Ours)}} \\
        \cmidrule(lr){1-6}
        \textbf{Model} & \textbf{Rel.} & \textbf{Str.} & \textbf{Grd.} & \textbf{Sim.} & \textbf{Avg} \\
        \midrule
        Debate-Feedback & 2.64 & 2.88 & 3.64 & 2.32 & 2.87 \\
        \textbf{OBJECTION} & \textbf{4.16} & \textbf{4.52} & \textbf{4.86} & \textbf{4.12} & \textbf{4.42} \\
        
        \midrule
        \midrule
        
        \multicolumn{6}{c}{\textbf{(B) Step-wise Analysis (Ours: Lawyer Agent)}} \\
        \cmidrule(lr){1-6}
        \textbf{Step} & \textbf{Rel.} & \textbf{Str.} & \textbf{Grd.} & \textbf{Sim.} & \textbf{Avg} \\
        \midrule
        Offense & \textbf{4.20} & 4.45 & 4.85 & \textbf{4.40} & 4.48 \\
        Unlawfulness & \textbf{4.20} & \textbf{4.87} & \textbf{5.00} & 4.20 & \textbf{4.57} \\
        Culpability & 4.07 & 4.27 & 4.73 & 3.67 & 4.19 \\
        \bottomrule
    \end{tabular}%
    }
\end{table}

\paragraph{Analysis and Insights}
As shown in Table \ref{tab:human_eval_compare} (A), the generic Debate-Feedback method fails to construct legally sound arguments, averaging a score of 2.87. It particularly struggles to mimic professional legal reasoning (Human Similarity: 2.32). In contrast, OBJECTION achieves an expert-level average of 4.42.

Furthermore, the step-wise analysis in (B) highlights that our Lawyer Agent excels particularly in the Unlawfulness stage (Avg 4.57, Grd 5.00), demonstrating exceptional capability in verifying strict legal requirements and establishing rigorous factual grounding without hallucination. Factual Grounding is also the highest-scoring criterion overall (4.86), indicating that the arguments stay anchored to the given facts rather than inventing exculpatory detail; the panel penalized any claim not traceable to the input. Grounding errors are nonetheless possible, and retrieval-augmented grounding over statutes and precedent is a natural mitigation we leave to future work.

\section{Conclusion}
We identified that the Guilty Bias stemming from prosecutorial narratives is a critical point in Legal NLP. To address this, we proposed OBJECTION, an inference-time adversarial pipeline. By combining a Lawyer Agent into the 3-step reasoning structure, our framework actively challenges the model's presumption of guilt.

Empirically, OBJECTION reduced the False Guilty Rate (FGR) on the Natural Innocent dataset, where fine-tuned SOTA models failed, while being robust to out-of-domain datasets without additional training.

Ultimately, this work implements the Presumption of Innocence at the system architecture level. Our pipeline favors preventing false convictions over raw accuracy. Nevertheless, OBJECTION attains the highest or tied-highest Macro-F1 in every backbone--dataset setting, maintaining a robust balance between guilty and innocent predictions. We present OBJECTION not as a simple technical improvement, but as a step toward normative alignment in Legal AI, assuring that digital judgments adhere to the basic principles of criminal justice.

\section{Limitations}
\label{limitations}
First, our pipeline is designed within the framework of the Civil Law system. Adaptation to Common Law systems, which focus on case law, demands further study. However, given that the adversarial principle (prosecution vs. defense) is a universal tenet of criminal justice, we argue that our Lawyer Agent's core mechanism—challenging narrative bias—remains valid across legal systems.

Second, our current Information Extraction (IE) module uses a generic SOAM (Subject-Object-Act-Mens Rea) schema to ensure scalability. Developing jurisdiction-specific schemas that represent the granular constitutive elements of individual crimes could further improve performance.

Third, introducing an adversarial agent inherently increases inference cost compared to single-pass baselines. However, our early exit mechanism mitigates this by terminating the process immediately upon finding clear innocence, achieving a lower expected token consumption than exhaustive debate methods (e.g., Debate-Feedback); see Appendix~\ref{sec:efficiency_detail} for more details.

Finally, we address only the binary guilty / not-guilty decision. Charge and sentencing prediction should exhibit the same prosecutorial bias, and our framework extends naturally---offense-stage arguments bear on charge severity, culpability-stage arguments on sentencing---which we leave as our primary follow-up.

\section{Ethical Considerations}
\label{ethics}
The power of criminal justice, including the judgment of guilt, is constitutionally delegated exclusively to authorized human experts and public officials. This study does not advocate for AI to usurp this sovereign power or make autonomous verdicts. Instead, we explore the potential of AI as a Decision Support System to assist legal professionals in detecting overlooked innocence and ensuring procedural fairness. Licenses for all external resources are listed in Appendix~\ref{sec:licenses}.

Concerning data privacy, all ``Natural Innocent'' judgments used in this study were collected from publicly available records released by the Supreme Court of Korea and have been strictly de-identified and anonymized to protect the privacy of all individuals involved. See Appendix~\ref{sec:dataset_detail} for detailed information.

Regarding the human expert evaluation detailed in Section~\ref{sec:human_eval}, the protocol involved active law enforcement investigators assessing the legal validity of generated arguments. This evaluation was conducted as a minimal-risk professional task assessment. It did not involve the collection of sensitive personal information from the participants, nor did it pose any physical or psychological risks. Residual failure modes of the pipeline, which motivate keeping a human in the loop, are analyzed in Appendix~\ref{sec:failure_modes}.

\section*{Acknowledgements}
This work was supported in part by the National Research Foundation of Korea (NRF) grant (RS-2023-00280883, RS-2023-00222663); by the National Research Foundation, Korea, under project BK21 FOUR(Dept. of Data Science, SNU, No. 5199990914569); by the Korea Institute of Science and Technology Information (KISTI) in 2026 (No. (KISTI)K26L3M1C1), aimed at developing KONI (KISTI Open Neural Intelligence), a large language model specialized in science and technology; and by the Institute of Information \& communications Technology Planning \& Evaluation (IITP) grant funded by the Korea government(MSIT) (RS-2025-02263754, Human-Centric Embodied AI Agents with Autonomous Decision-Making); by grant (25202MFDS003) from Ministry of Food and Drug Safety in 2025; by AI-BIO Research Grant through Seoul National University; Institute of Information \& communications Technology Planning \& Evaluation (IITP) grant funded by the Korea government(MSIT) (No. RS-2025-25442149, LG AI STAR Talent Development Program for Leading Large-Scale Generative AI Models in the Physical AI Domain).

\bibliography{custom}
\newpage

\appendix

\section{Prompt Templates}\label{sec:appendix}

\begin{promptbox}[prompt:soam]{SOAM Schema Extraction}
\textbf{[System]} \\
You are a legal information extraction assistant. Your task is to extract SOAM elements only from the given facts. Do not fabricate or infer missing information. Output a strict JSON object only, without any other text.

\vspace{1em} 

\textbf{[User]} \\
<ROLE> \\
You are a SOAM element extractor. \\
</ROLE> \\
<RULES> \\
1) Extract only from the facts; if missing, fill with an empty string. \\
2) Each slot must contain two fields: value and evidence (or just value). \\
3) Output only a JSON object, the top-level keys must be: Act, MensRea, Object, Subject \\
</RULES> \\
<INPUT> \\
\{case\_text\} \\
</INPUT> \\
<OUTPUT\_JSON\_EXAMPLE> \\
\{\{ \\
"Act": \{\{"value": "...", "evidence": "..."\}\}, \\
"MensRea": \{\{"value": "...", "evidence": "..."\}\}, \\
"Object": \{\{"value": "...", "evidence": "..."\}\}, \\
"Subject": \{\{"value": "...", "evidence": "..."\}\} \\
\}\} \\
</OUTPUT\_JSON\_EXAMPLE> \\
\end{promptbox}

\paragraph{SOAM Extraction Prompt}
{Prompt}~\ref{prompt:soam} shows the zero-shot template used to extract the Subject, Object, Act, and Mens Rea elements from the case description, together with the supporting evidence span for each element.

\begin{promptbox}[prompt:judge1]{Initial Judgement Prompt}
\textbf{[System]} \\
You are a criminal law expert. Please strictly output the specified LJP block as required. Do not output article numbers, case numbers, or superfluous text irrelevant to the format.

\vspace{1em} 

\textbf{[User]} \\
<CASE\_FACTS> \\
\{case\_text\} \\
</CASE\_FACTS> \\
\{ie\_block\} \\
<QUESTION> \\
Based solely on the facts above and IE elements, please determine whether the conduct meets the Constituent Elements of the crime (Offense). \\
{[Scope of Judgment]} \\
1. **Determine Only**: Whether the conduct meets the objective and subjective constituent elements defined by criminal law.\\
2. **Strictly Forbidden**: Considering any justification grounds (e.g., Self-Defense) or excuse grounds (e.g., Insanity).\\
3. Even if justification/excuse grounds exist, if the constituent elements are met, you must label **YES** for this stage.\\
</QUESTION>\\
{[Offense]} \\
- label: YES or NO \\
- grounds:\\
1) ...\\
2) ...\\
\end{promptbox}

\paragraph{Initial Judgment Prompt}
{Prompt}~\ref{prompt:judge1} shows the prompt template used for the Initial Judgment (Judge 1).
This template is applied across the Offense, Unlawfulness, and Culpability stages.
The \textit{Question} block explicitly adapts its instructions for each stage:
Checking constituent elements for Offense, evaluating justification grounds (e.g., self-defense) for Unlawfulness, and determining the responsibility capacity (e.g., insanity) for Culpability, ensuring that the model focuses strictly on the legally relevant criteria of the current stage.

\begin{promptbox}[prompt:def1]{Defense(Offense)}
\textbf{[User]} \\
\{case\_text\}\\
</CASE\_FACTS>\\
\{ie\_block\}\\
{[Offense Initial Judgment]}\\
\{judge\_block\}\\
<QUESTION>\\
You are a Defense Lawyer. Please read the [Offense Initial Judgment] and write a defense opinion for the defendant.\\
Goal: Find mitigating circumstances such as "Constituent Elements Not Met (Offense NO)" or "Attempt/Discontinuation".\\
Strategy (In Dubio Pro Reo):\\
1. **Challenge Elements**: Raise factual or legal doubts regarding the nature of the act, causation, or subjective elements (Intent/Negligence).\\
2. **Attempt \& Discontinuation**: If the act was not completed or voluntarily abandoned, point this out.\\
3. **Targeted Rebuttal**: You must directly cite original text from the case facts (Evidence) to support your rebuttal.\\
{[Stage Scope]}\\
1. **Defend Only**: Issues regarding the applicability of Constituent Elements.\\
2. **Strictly Forbidden**: Mentioning Self-Defense, Necessity, Insanity, Age, etc., which belong to subsequent stages.\\
3. If the case only involves the above subsequent issues, output "No Objection" for this stage.\\
Please output "Defense Opinion (Defense-Offense)". If the prosecution's judgment is unassailable (or points belong to later stages), output "No Objection".\\
</QUESTION>\\
{[Defense-Offense]}\\
- point1: ...

\end{promptbox}

\paragraph{Defense at the Offense Stage}
Prompt~\ref{prompt:def1} instructs the Lawyer to challenge the establishment
of constituent elements or argue for attempt, discontinuation, or reasonable doubt under
\textit{In Dubio Pro Reo}, while sharing the same System Prompt as the Initial Judge.

\begin{promptbox}[prompt:def2]{Defense(Unlawfulness)}
\textbf{[User]} \\
<CASE\_FACTS>\\
\{case\_text\}\\
</CASE\_FACTS>\\
\{ie\_block\}\\
{[Illegality Initial Judgment]}\\
\{judge\_block\}\\
<QUESTION>\\
You are a Defense Lawyer. Please read the [Illegality Initial Judgment] and write a defense opinion for the defendant.\\
Goal: Argue for Self-Defense, Necessity, or other Justification grounds.\\
Strategy:\\
1. **Exploit Favorable Circumstances**: Focus on descriptions like "Victim's fault", "Imminent danger", "Unavoidable".\\
2. **Challenge Strict Standards**: If the prosecution sets the bar too high for Defense/Necessity, argue that the urgency of the situation should be considered.\\
{[Stage Scope]}\\
1. **Defend Only**: Justification grounds such as Self-Defense, Necessity, Justified Act.\\
2. **Strictly Forbidden**: Issues of Responsibility such as Insanity, Age (leave for Culpability).\\
3. If the case involves only responsibility issues, output "No Objection" for this stage.\\
Please output "Defense Opinion (Defense-Illegality)". If there are clearly no justification grounds (or they belong to responsibility issues), output "No Objection".\\
</QUESTION>\\
{[Defense-Illegality]}\\
- point1: ...

\end{promptbox}

\paragraph{Defense at the Unlawfulness Stage}
Prompt~\ref{prompt:def2} directs the Lawyer to identify justification grounds 
(e.g., self-defense or necessity) and to argue for justification even in legally ambiguous
situations.

\begin{promptbox}[prompt:def3]{Defense(Culpability)}
\textbf{[User]} \\
<CASE\_FACTS> \\
\{case\_text\} \\
</CASE\_FACTS> \\
\{ie\_block\}\\
{[Culpability Initial Judgment]}\\
\{judge\_block\}\\
<QUESTION>\\
You are a Defense Lawyer. Please read the [Culpability Initial Judgment] and write a defense opinion for the defendant.\\
Goal: Argue for lack of Responsibility Capacity (e.g., Insanity, Minor) or lack of Expectability.\\
Strategy:\\
1. **Highlight Special States**: Focus on Age, History of Mental Illness, Intoxication (if pathological), Duress, etc.\\
2. **Humanitarian Perspective**: If the defendant was in a desperate situation, argue the inevitability of their conduct.\\
{[Stage Scope]}\\
1. **Defend Only**: Criminal Responsibility Capacity (Insanity, Age, Expectability).\\
Please output "Defense Opinion (Defense-Culp)". If there are no relevant circumstances, output "No Objection".\\
</QUESTION>\\
{[Defense-Culp]}\\
- point1: ...

\end{promptbox}

\paragraph{Defense at the Culpability Stage}
Prompt~\ref{prompt:def3} prompts the Lawyer to argue for a lack of responsibility
by highlighting special states such as insanity, minor age, or lack of expectability.

\begin{promptbox}[prompt:judge2]{Final Judgement Aggregation}
\textbf{[System]} \\
You are a criminal law expert with 15 years of experience. You should make strict, restrained, and verifiable judgments based on facts. \\
You will see the Prosecution's Initial Judgment and the Defense Points. Adopt only those defense points that are consistent with the facts, and provide a final judgment.\\
<CORE\_RULES>\\
1) Judge based ONLY on the facts appearing in <CASE\_FACTS>.\\
2) Adopt ONLY defense points consistent with the facts; ignore any points beyond the facts.\\
3) Output must strictly follow the specified format: label + grounds.\\
</CORE\_RULES>\\

\vspace{1em} 

\textbf{[User]} \\
<CASE\_FACTS> \\
\{case\_text\}\\
</CASE\_FACTS>\\
\{ie\_block\}\\
{[Prosecution - Initial Judgment]}\\
\{judge\_initial\}\\
{[Defense Opinion]}\\
\{lawyer\_defense\}\\
<QUESTION>\\
Adopt only defense points consistent with the facts, and provide the final judgment for \{block\_name\}.\\
</QUESTION>\\
{[\{out\_block\}]} \# stage name\\
- label: \{label\_line\}\\
- grounds:\\
1) ...\\

\end{promptbox}

\paragraph{Final Judgment Aggregation}
Prompt~\ref{prompt:judge2} aggregates the Initial Judgment and the Lawyer’s defense arguments
and applies rigorous fact-checking to produce the final verdict for the current stage.

\section{About ``Natural Innocent'' Dataset} \label{sec:dataset_detail}
\subsection{Construction Pipeline}
The \textbf{Natural Innocent} dataset is constructed from criminal judgments publicly released by the South Korean government, specifically from the \textit{Korean Law Information Center} and the courts. All government judgments undergo a strict anonymization process to protect personal privacy.

To ensure label integrity, Class 1 (Not Guilty) contains only cases with a final verdict of \textbf{Not Guilty (Acquittal)}, while Class 0 (Guilty) contains cases with a final guilty verdict, added to enable guilty-side metrics and balanced evaluation. Furthermore, as securing the detailed description of facts (prosecutorial charges) is central to our input construction, we restricted our scope to \textbf{first-instance criminal judgments}. In the South Korean legal system, the first instance serves as the fact-finding trial (\textit{Tatfrage}), whereas the second and third instances focus on legal interpretation (\textit{Rechtsfrage}). Therefore, first-instance judgments provide the most in-depth description of factual relationships and the court's reasoning regarding the charges.

From the pool of first-instance acquittals, we filtered cases using \textbf{keyword matching} and categorized them into the three stages of the trichotomous framework. We identified whether the court's reasoning contained specific keywords or adjudicated specific legal issues relevant to each stage:

\begin{itemize}
    \item \textbf{Offense-level Acquittals:} We primarily targeted the idiomatic judicial phrase ``\textit{lack of proof of crime}'' (insufficiency of evidence). We also collected cases based on specific contentious issues inherent to individual crimes, such as distinguishing fraud from simple \textbf{debt default}, the \textbf{public interest} aspect in defamation, or determining whether failing to return \textbf{mistakenly remitted funds} constitutes embezzlement.
    \item \textbf{Unlawfulness-level Acquittals:} We collected cases containing keywords related to justification grounds, including ``\textit{Excessive Defense},'' ``\textit{Emergency Evacuation},'' ``\textit{Justifiable Act},'' ``\textit{Self-Defense},'' and ``\textit{Victim's Consent}.''
    \item \textbf{Culpability-level Acquittals:} We used keywords related to responsibility and attribution, such as ``\textit{Theory of Expectability},'' ``\textit{Capacity to Discern},'' ``\textit{Criminal Responsibility},'' and ``\textit{Mistake of Law}.''
\end{itemize}

The specific keywords and their distribution are detailed in Table \ref{tab:keyword_stats}. All resulting stage assignments were verified by a police investigator.

\begin{table}[!h]
\centering
\small
\resizebox{\columnwidth}{!}{%
\begin{tabular}{l l r}
\toprule
\textbf{Category} & \textbf{Keyword (Translation)} & \textbf{Count} \\
\midrule
\multirow{10}{*}{\textbf{Offense}} 
 & Debt Default (Fraud) & 192 \\
 & Lack of Proof & 189 \\
 & Property of Others & 187 \\
 & Public Interest (Defamation) & 186 \\
 & Negation of Causality & 151 \\
 & Simple Debt Default & 106 \\
 & Return (Embezzlement) & 102 \\
 & Potential Loss & 92 \\
 & Custodian Status & 79 \\
 & Pecuniary Loss & 68 \\
\midrule
\multirow{7}{*}{\textbf{Unlawfulness}} 
 & Not against Social Rules & 175 \\
 & Self-Defense & 118 \\
 & Victim's Consent & 51 \\
 & Justifiable Act & 49 \\
 & Emergency Evacuation & 20 \\
 & Excessive Defense & 10 \\
 & Act due to Duty & 8 \\
\midrule
\multirow{9}{*}{\textbf{Culpability}} 
 & No Intent & 87 \\
 & Mistake of Justification & 29 \\
 & Capacity to Decide & 21 \\
 & Awareness of Illegality & 20 \\
 & Capacity to Discern & 20 \\
 & Lawful Act (Expectability) & 19 \\
 & Expectability & 15 \\
 & Mistake of Law & 5 \\
 & Medical Treatment & 5 \\
\bottomrule
\end{tabular}%
}
\caption{Detailed statistics of keyword-based filtering for the Natural Innocent dataset. Keywords are categorized into the three stages of the trichotomous reasoning framework. Top keywords only for Offense and Culpability.}
\label{tab:keyword_stats}
\end{table}
\subsection{Overview \& Statistics}
This section summarizes the dataset: its provenance and scale (Table~\ref{tab:nat_overview}), the distribution of verdict types under the trichotomous framework (Table~\ref{tab:verdict_dist}), the ten most frequent charges (Table~\ref{tab:top10_charges}), and the token length of case descriptions (Table~\ref{tab:token_stats}).

\begin{table}[h]
\centering
\small
\begin{tabular}{ll}
\hline
\textbf{Item} & \textbf{Description} \\
\hline
Source & Real-world Korean criminal judgments \\
Verdict Type & Guilty \& Not Guilty (Acquittal) \\
Total Cases & 3,412 \\
Jurisdiction & South Korea \\
Crime Domains & Fraud, Violence, Narcotics, etc. \\
Purpose & Balanced FGR/FNR evaluation \\
\hline
\end{tabular}
\caption{Overview of the Natural Innocent dataset.}
\label{tab:nat_overview}
\end{table}

\begin{table}[h!]
\centering
\small
\begin{tabular}{l l r r}
\hline
\textbf{Class} & \textbf{Legal Ground} & \textbf{Count} & \textbf{Ratio (\%)} \\
\hline
Class 0 & Guilty & 1,291 & 37.8 \\
Class 1 & Offense Not Established & 1,464 & 42.9 \\
Class 2 & Justification & 431 & 12.6 \\
Class 3 & No Culpability & 226 & 6.6 \\
\hline
Total &  & 3,412 & 100.0 \\
\hline
\end{tabular}
\caption{Distribution of verdict types under the trichotomous framework (Offense--Unlawfulness--Culpability).}
\label{tab:verdict_dist}
\end{table}

\begin{table}[h!]
\centering
\small
\begin{tabular}{r l r r}
\hline
\textbf{Rank} & \textbf{Charge} & \textbf{Count} & \textbf{Ratio (\%)} \\
\hline
1 & Fraud & 657 & 19.3 \\
2 & Assault & 249 & 7.3 \\
3 & Aggravated Fraud & 241 & 7.1 \\
4 & Embezzlement & 163 & 4.8 \\
5 & Agg.\ Assault / Injury & 129 & 3.8 \\
6 & Occupational Embezzlement & 111 & 3.3 \\
7 & Property Damage & 110 & 3.2 \\
8 & Injury & 109 & 3.2 \\
9 & Defamation & 100 & 2.9 \\
10 & Cyber Fraud & 98 & 2.9 \\
\hline
\end{tabular}
\caption{Top-10 offense distribution in the Natural Innocent dataset. All charges are normalized to offense-level categories and reported in English.}
\label{tab:top10_charges}
\end{table}

\begin{table}[h!]
\small
\centering
\begin{tabular}{l r}
\hline
\textbf{Statistic} & \textbf{Tokens} \\
\hline
Tokenizer & Qwen-2.5-7B \\
Mean & 781 \\
Median & 575 \\
25th Percentile & 301 \\
75th Percentile & 1,030 \\
IQR & 301 -- 1,030 \\
\hline
\end{tabular}
\caption{Token-length statistics of case descriptions in the Natural Innocent dataset.}
\label{tab:token_stats}
\end{table}

\subsection{Synthetic vs. Natural Innocent}
\begin{figure*}[h!]
    \centering
    \includegraphics[width=0.8\textwidth]{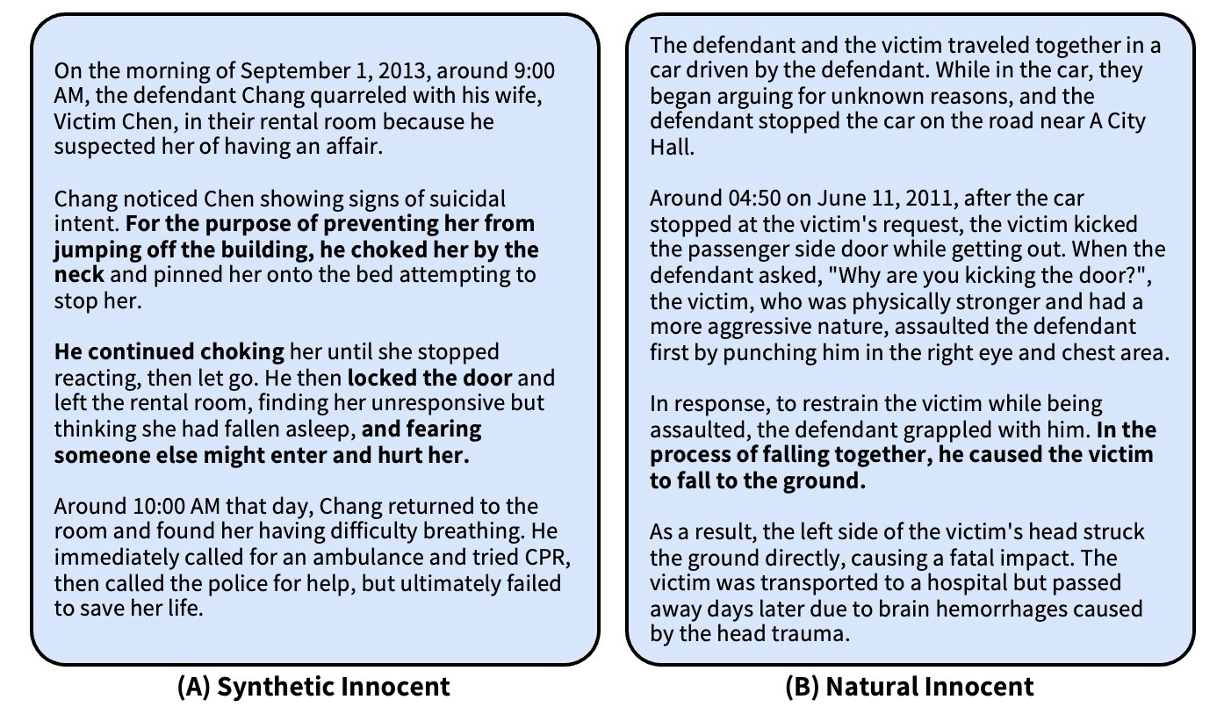}
    \caption{Comparison of Synthetic vs. Natural Innocent narratives. \textbf{(Left)} Synthetic cases usually contain forced, \textbf{implausible scenarios with explicit exculpatory cues}, risking shortcut learning. \textbf{(Right)} Natural Innocent cases present realistic, logical narratives where innocence must be \textit{inferred} from implicit contextual details (e.g., incidental injury during self-defense), requiring genuine legal reasoning.}
    \label{fig:synnat}
\end{figure*}
Figure~\ref{fig:synnat} shows the difference between Synthetic Innocent and Natural Innocent.

\paragraph{Synthetic Innocent (Left)}
The synthetic innocent case shows clear evidence of coerced counterfactual rewriting. Although preventing a suicidal act could plausibly justify physical restraint, the narrative escalates into implausible behavior—continuing to choke until the victim becomes unresponsive and then locking the door out of fear that someone else might enter. These actions break common-sense causal and behavioral expectations and read as overtly injected exculpatory cues rather than organically arising facts. Such unnatural constructions make innocence highly salient and explicit, which risks shortcut learning during fine-tuning: models may learn to associate conspicuous, unrealistic signals with acquittal rather than develop genuine legal reasoning.

\paragraph{Natural Innocent (Right)}
The natural innocent case presents a temporally coherent, realistic narrative in which innocence must be inferred rather than stated. The victim initiates the assault; the defendant attempts to restrain him; and the fatal injury occurs incidentally as both fall, without any explicit claim about intent. Exculpatory clues—self-defensive context, proportional response, and accidental outcome—are embedded implicitly in the sequence of events. This forces models to integrate context and causality to infer non-culpability, making the case substantially harder and more faithful to real-world acquittals than synthetic innocence with explicit cues.

\section{Detailed Case Study (Qualitative Analysis)}
\label{abl:case_study}

\begin{figure}[h]
\centering
\includegraphics[width=\linewidth]{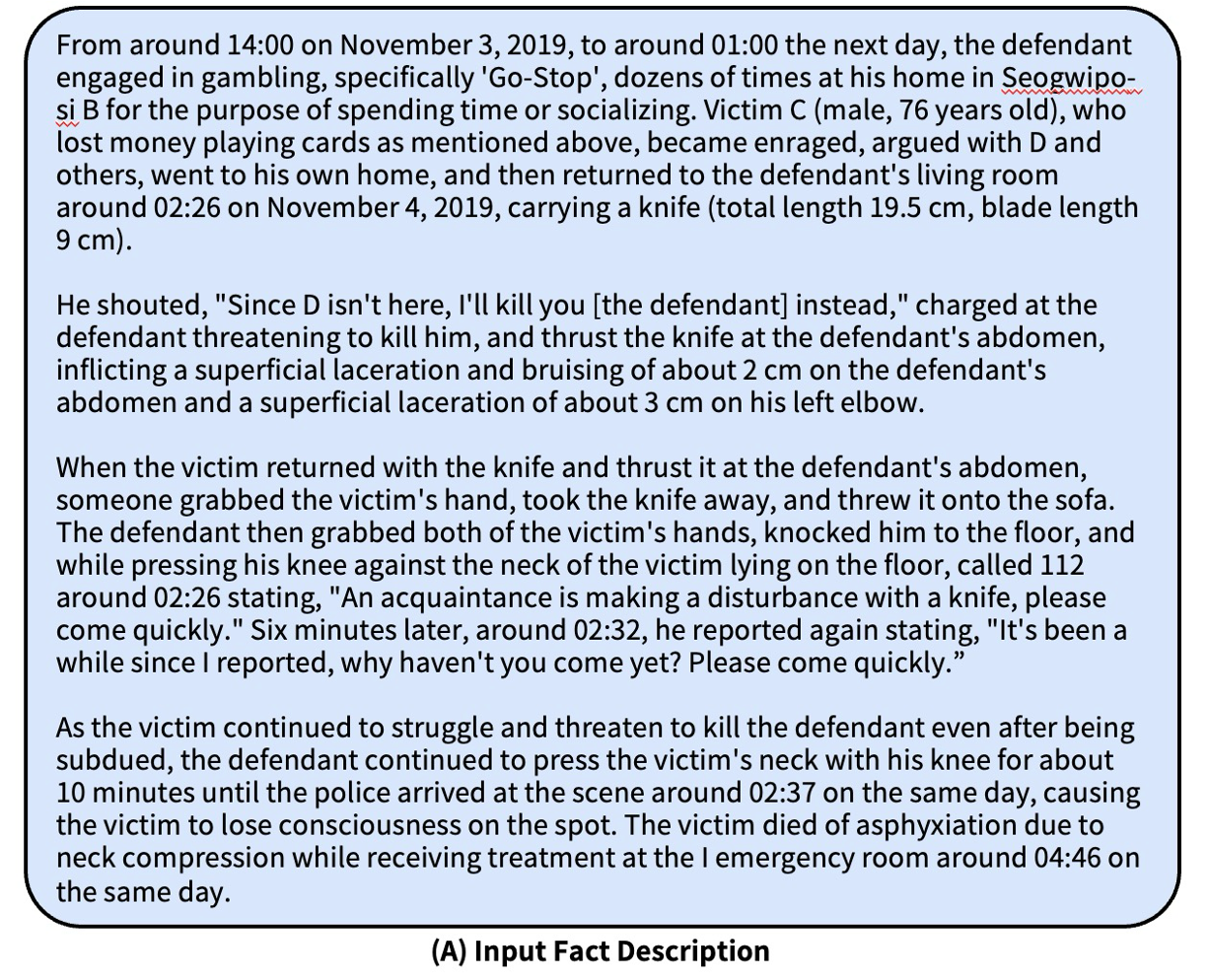}
\caption{Input fact description of the case involving an armed attack and subsequent restraint.}
\label{fig:case-input}
\end{figure}

\paragraph{Input Facts.}
Figure~\ref{fig:case-input} shows the input fact description of a real-world criminal case.
Following a gambling-related dispute, the victim returned to the defendant’s residence armed
with a knife and explicitly threatened to kill him. The victim initiated the attack and stabbed
the defendant, after which the knife was removed, and the defendant attempted to restrain the
victim while repeatedly calling the police. Despite being subdued, the victim continued to
struggle and issue death threats, and later died due to neck compression while being restrained
until police arrival.

\begin{figure}[h]
\centering
\includegraphics[width=\linewidth]{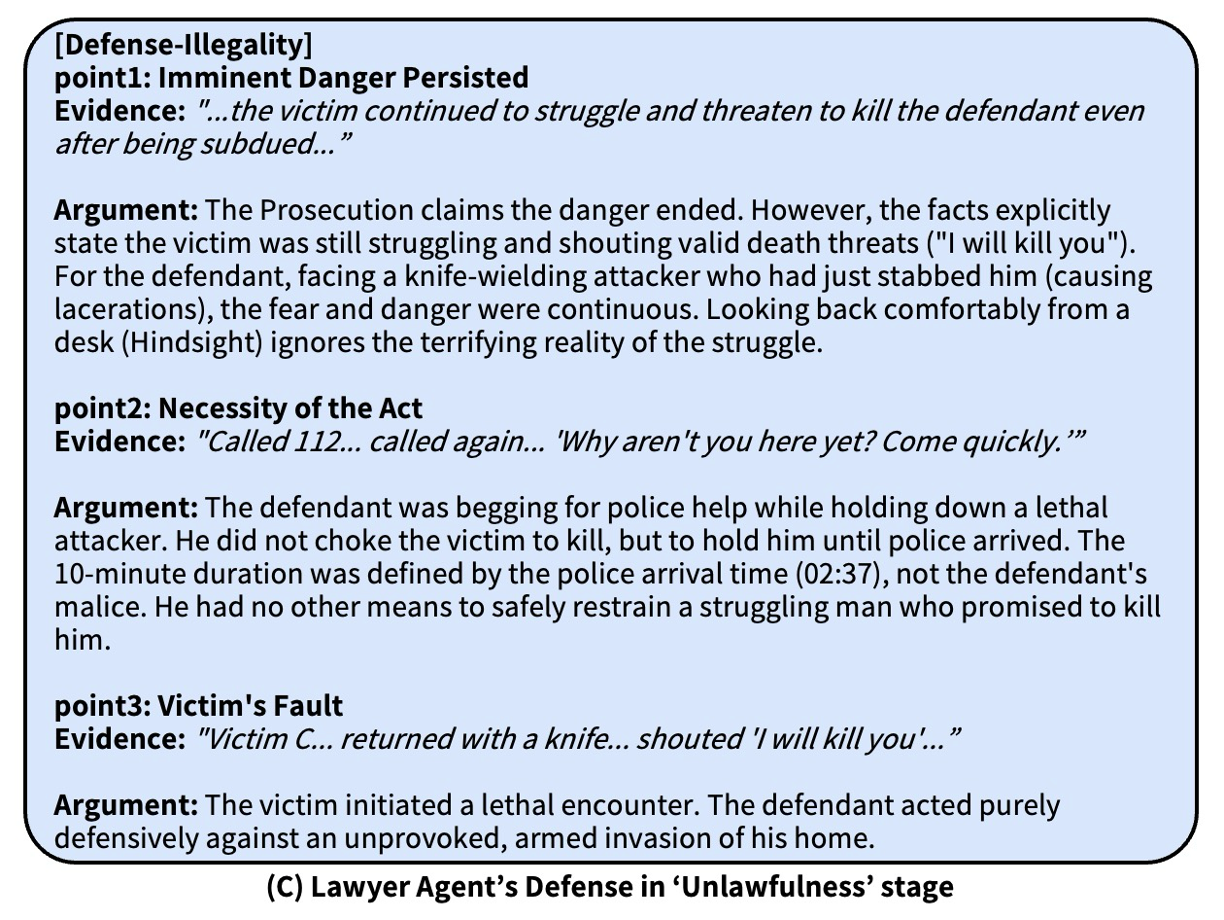}
\caption{Lawyer Agent’s stage-specific defense arguments highlighting imminence, necessity, and victim fault.}
\label{fig:case-lawyer-defense}
\end{figure}

\paragraph{SOAM Schema Extraction.}
To provide operational clarity on our schema and address potential conceptual ambiguity, we demonstrate the step-by-step mapping of the narrative shown in Figure~\ref{fig:case-input}.
Our extraction module maps the unstructured facts into the following elements:
\begin{itemize}[topsep=5pt, parsep=0pt, itemsep=0.3pt]
    \item \textbf{Subject ($S$):} The defendant.
    \item \textbf{Object ($O$):} Victim C.
    \item \textbf{Actus Reus ($A$):} The defendant ``grabbed both of the victim's hands, knocked him to the floor, and continued to press the victim's neck with his knee for about 10 minutes.'' (This captures strictly the objective physical conduct and the resulting death).
    \item \textbf{Mens Rea ($M$):} The defendant ``called 112 around 02:26... [and] reported again stating, 'It's been a while since I reported, why haven't you come yet?'.'' (This captures the subjective state: the intent was to restrain the attacker until help arrived, not a malicious intent to kill).
\end{itemize}
\begin{table*}[t]
\centering
\small
\begin{tabular}{l l c c c c}
\toprule
\textbf{Backbone} & \textbf{Method} & \textbf{Avg Calls} & \textbf{Input Tok.} & \textbf{Output Tok.} & \textbf{Latency (s)} \\
\midrule
 \multirow{4}{*}{\textbf{Qwen-2.5-7B}}
 & Baseline D & 4.0 & 1,830 & 1,392 & \textbf{3.35} \\
 & Self-Refine & 4.0 & 2,724 & 2,952 & 4.86 \\
 & Debate & 8.0 & 16,167 & 12,997 & 16.08 \\
 & \textbf{OBJECTION} & 7.17 & 6,149 & 3,285 & \textbf{6.77} \\
\midrule
\multirow{4}{*}{\textbf{Llama-3.1-8B}}
 & Baseline D & 4.0 & 1,830 & 637 & \textbf{3.63} \\
 & Self-Refine & 4.0 & 2,724 & 1,351 & 5.45 \\
 & Debate & 8.0 & 9,151 & 5,947 & 18.70 \\
 & \textbf{OBJECTION} & 5.57 & 3,717 & 1,168 & \textbf{5.77} \\
\bottomrule
\end{tabular}
\caption{Efficiency comparison on the CAIL dataset. \textbf{Avg Calls}, \textbf{Tokens}, and \textbf{Latency} for OBJECTION are weighted averages representing the early-exit distribution. OBJECTION is significantly faster than the multi-agent Debate baseline, thanks to fewer expected interactions.}
\label{tab:efficiency}
\end{table*}
In real-world criminal adjudications, \textit{Mens Rea} is rarely stated explicitly. Instead, it is inextricably embedded within the context.
As demonstrated above, our pipeline handles this edge case by instructing the extraction module to capture contextual actions (e.g., calling the police twice) that implicitly reveal the subjective state.
By isolating these inferred subjective elements ($M$) from the fatal physical outcome ($A$), the SOAM schema prevents the model from blindly equating a lethal consequence with malicious intent.
This structured breakdown provides the exact factual anchor the Lawyer Agent needs to construct the defense arguments later shown in Figure~\ref{fig:case-lawyer-defense}.

\begin{figure}[h]
\centering
\includegraphics[width=\linewidth]{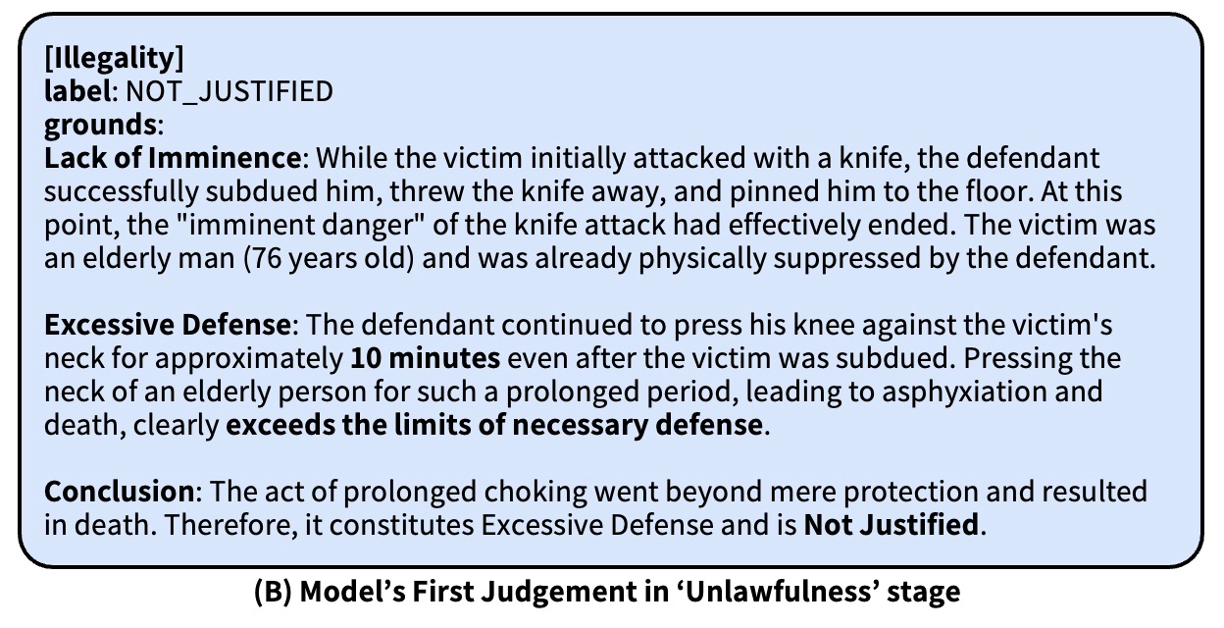}
\caption{Model’s initial judgment at the Unlawfulness stage, concluding that the act was not justified.}
\label{fig:case-initial-unlawfulness}
\end{figure}
\paragraph{Initial Judgment.}
As shown in Figure~\ref{fig:case-initial-unlawfulness}, the base model initially judged the
defendant’s actions as \textit{Not Justified} at the Unlawfulness stage. The model reasoned that
the imminent danger had ended once the knife was removed and that maintaining neck compression
for an extended period constituted excessive defense.

\paragraph{Lawyer Agent Intervention.}
Figure~\ref{fig:case-lawyer-defense} illustrates the Lawyer Agent’s intervention at the
Unlawfulness stage. Without presenting new facts, the agent reorganizes existing evidence into
legally salient defense arguments, emphasizing the persistence of imminent danger, the necessity
of restraint while awaiting a delayed police response, and the victim’s fault as the initiator of
a lethal armed attack.

\begin{figure}[h]
\centering
\includegraphics[width=\linewidth]{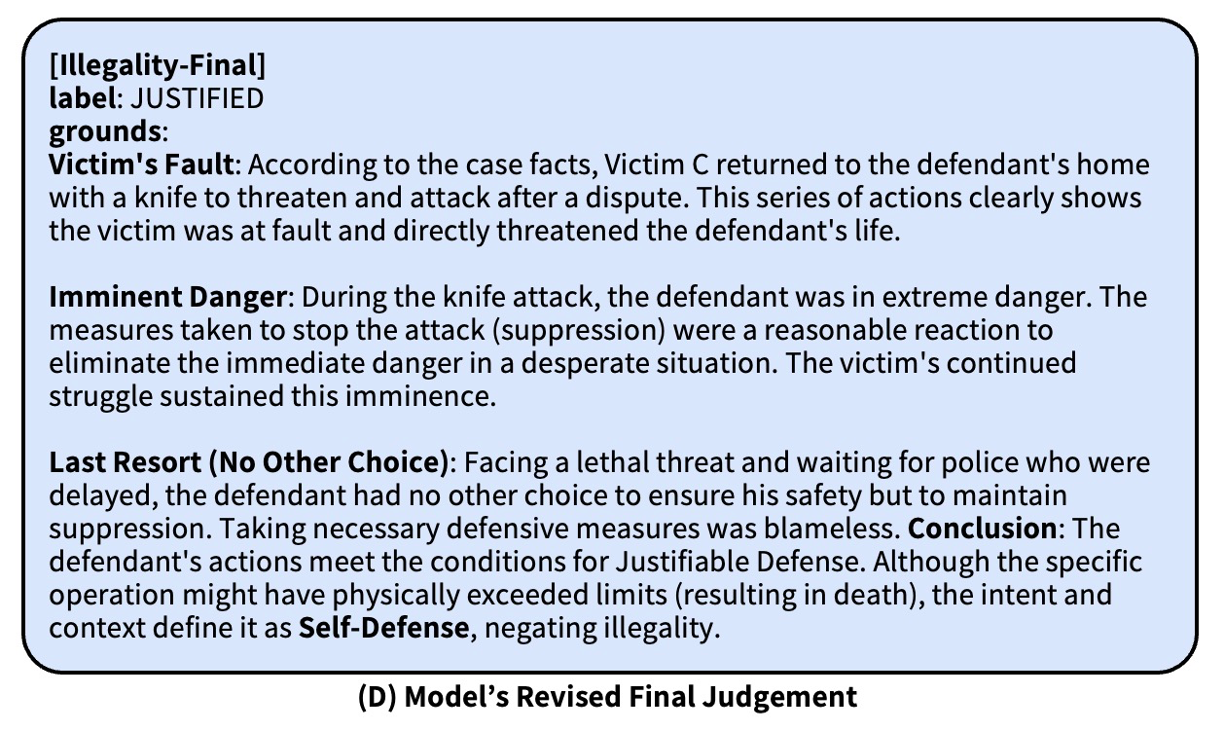}
\caption{Model’s revised final judgment accepting justification after defense intervention.}
\label{fig:case-revised-judgment}
\end{figure}
\paragraph{Revised Judgment.}
After incorporating the Lawyer Agent’s defense arguments, the model revises its judgment, as
shown in Figure~\ref{fig:case-revised-judgment}, concluding that the defendant’s conduct was
\textit{Justified}. The revised decision recognizes continued imminence, lack of reasonable
alternatives, and last-resort necessity, thereby negating unlawfulness through justifiable
self-defense.

\section{Additional Implementation Details}
\paragraph{Backbone Large Language Models}
We employed the following open-weight large language models in our experiments. The primary backbone model was \texttt{Qwen2.5-7B-Instruct}  (Alibaba Cloud). For robustness, we additionally evaluated \texttt{Llama-3.1-8B-Instruct} (Meta) and \texttt{Gemma2-9B-it} (Google).

\paragraph{Inference Infrastructure}
All inference experiments were conducted using \texttt{vLLM} (v0.11.0) for
high-throughput model serving via an OpenAI-compatible API server.
Experiments were run on a workstation equipped with a single
NVIDIA RTX PRO 6000 Blackwell GPU with 96GB VRAM.
Models were loaded with automatic precision (typically \texttt{bfloat16})
to balance numerical stability and throughput.

\paragraph{Inference Hyperparameters}
To guarantee equitable comparison across all pipelines, we used consistent generation
settings throughout the experiments:
Temperature was set to 0.0 (greedy decoding for reproducibility),
The maximum number of generation tokens was 2,048,
The maximum context length was 8,192 tokens,
and GPU memory utilization was capped at 0.90 to reserve capacity
for the vLLM key--value cache.

\paragraph{LJPIV Fine-tuning Baseline Implementation}
For the LJPIV baseline, we fine-tuned \texttt{Qwen/Qwen2.5-7B-Instruct} using
LoRA (Low-Rank Adaptation).
Fine-tuning was performed in a separate environment equipped with a single
NVIDIA RTX A6000 GPU (48GB VRAM).

\paragraph{Fine-tuning Configuration}
We used the PEFT library (v0.17.1) with the LoRA method.
LoRA was applied to the following target modules:
\texttt{q\_proj}, \texttt{k\_proj}, \texttt{v\_proj}, \texttt{o\_proj},
\texttt{gate\_proj}, \texttt{up\_proj}, and \texttt{down\_proj}.
The LoRA rank was set to 8, with LoRA alpha set to 32 and dropout set to 0.1.

\paragraph{Training Hyperparameters}
Optimization was performed using AdamW (PyTorch implementation)
with a learning rate of $5 \times 10^{-5}$ and a cosine learning rate scheduler.
The batch size was 16 per device with gradient accumulation steps set to 2,
resulting in an effective batch size of 32.
Models were trained for three epochs using \texttt{bfloat16} precision,
with a maximum sequence length of 2,048 tokens.
\section{Inference Efficiency Analysis} \label{sec:efficiency_detail}
To evaluate the computational practicality of our framework, we measured the average processing time (latency) and token consumption per case on the \textbf{CAIL} dataset. Experiments were conducted on a single \texttt{NVIDIA RTX PRO 6000 blackwell} GPU using the \texttt{vLLM} backend.

We compared \textbf{OBJECTION} against three baselines: Baseline D (Chain-of-Thought), Self-Refine, and Debate (Multi-Agent). Table \ref{tab:efficiency} summarizes the results across two backbones.

\paragraph{Efficiency Gains via Early Exit}
As shown in Table \ref{tab:efficiency}, the \textbf{Debate} baseline requires a fixed number of interactions ($8.0$ calls), leading to consistently high latency. In contrast, \textbf{OBJECTION} utilizes an early-exit mechanism, substantially decreasing the expected computational cost based on the verdict distribution:
\begin{itemize}
    \item \textbf{Llama-3.1:} Since the model frequently dismisses charges at the Offense stage ($62.68\%$), the average API calls drop to \textbf{5.57}, shortening latency to \textbf{5.77s}.
    \item \textbf{Qwen-2.5:} With a higher conviction rate, the model averages \textbf{7.17} calls with a latency of \textbf{6.77s}.
\end{itemize}

This dynamic computation enables \textbf{OBJECTION} to achieve latency reductions of approximately \textbf{69.1\%} (Llama) and \textbf{57.9\%} (Qwen) compared to the Debate baseline. This shows that our structured agent system offers a superior efficiency-performance trade-off compared to unstructured debate frameworks.

\newpage

\section{Ablation on LJPIV-CAIL dataset} \label{sec:cail_ablation}
Table~\ref{tab:ablation_res2} provides the full ablation study results on the LJPIV-CAIL benchmark, supplementing the main analysis in Section~\ref{sec:analysis}. The results follow the same trend observed in the ELAM dataset, confirming the integral role of our Lawyer Agent.

\begin{table}[h]
\centering
\small
\caption{Ablation Results (w/ \texttt{Qwen 2.5}) on LJPIV-CAIL dataset. The Lawyer Agent is effective in a low False Guilty Rate (FGR).}
\label{tab:ablation_res2}
\resizebox{\columnwidth}{!}{%
\begin{tabular}{lcccc}
\toprule
\textbf{Configuration} & \textbf{G-F1} & \textbf{NG-F1} & \textbf{FGR}$\downarrow$ & \textbf{FNR}$\downarrow$ \\
\midrule
\textbf{OBJECTION (Full)} & \textbf{0.75} & \textbf{0.80} & \textbf{10.71} & 33.93 \\
(-) Schema & 0.68 & 0.76 & 13.21 & 42.14 \\
(-) Lawyer Agent & 0.66 & 0.69 & 27.86 & 37.14 \\
(-) 3-Step Structure & 0.35 & 0.57 & 31.07 & 72.14 \\
Ours w/ Normal Critic & \textbf{0.75} & 0.59 & 55.71 & \textbf{5.71} \\
\midrule
LJPIV & 0.80 & 0.83 & 9.29 & 27.14 \\
LJPIV + Lawyer Agent & 0.02 & 0.67 & 0.00 & 98.93 \\
\bottomrule
\end{tabular}%
}
\end{table}

\section{Statistical Significance Testing}
\label{sec:significance}

\paragraph{Protocol.}
All experiments use greedy decoding ($T=0.0$), so the results carry no sampling variance and run-level variance testing is inapplicable. For every backbone--dataset setting we apply an exact McNemar test to the paired verdicts on the innocent subset, which tests FGR directly. All comparisons use the same per-case predictions reported in Table~\ref{tab:main_results_unified}.

\begin{table*}[t]
\centering
\footnotesize
\setlength{\tabcolsep}{3pt}
\caption{Exact McNemar tests on FGR. Each cell gives $\Delta$FGR (OBJECTION minus the baseline, in points; negative favors OBJECTION) and the $p$-value. \textbf{Bold} marks a significant advantage for OBJECTION at $\alpha=.05$.}
\label{tab:significance_main}
\begin{tabular}{@{}l r@{\ \ }l r@{\ \ }l r@{\ \ }l@{}}
\toprule
\multirow{2}{*}{\textbf{Setting}} & \multicolumn{2}{c}{\textbf{vs. Baseline D}} & \multicolumn{2}{c}{\textbf{vs. Self-Refine}} & \multicolumn{2}{c}{\textbf{vs. Debate}} \\
\cmidrule(lr){2-3} \cmidrule(lr){4-5} \cmidrule(lr){6-7}
 & $\Delta$ & $p$ & $\Delta$ & $p$ & $\Delta$ & $p$ \\
\midrule
CAIL / Qwen    & \textbf{$-$42.5} & \textbf{9.2e$-$35} & \textbf{$-$38.6} & \textbf{1.1e$-$27} & \textbf{$-$11.4} & \textbf{.018} \\
CAIL / Llama   & \textbf{$-$20.0} & \textbf{2.3e$-$10} & \textbf{$-$28.6} & \textbf{2.7e$-$11} & $+$20.0 & 7.5e$-$15 \\
CAIL / Gemma   & \textbf{$-$31.1} & \textbf{3.4e$-$24} & \textbf{$-$33.6} & \textbf{1.7e$-$24} & $+$3.9 & .215 \\
\cmidrule(lr){1-7}
ELAM / Qwen    & \textbf{$-$33.2} & \textbf{4.4e$-$24} & \textbf{$-$34.4} & \textbf{9.4e$-$20} & \textbf{$-$3.8} & \textbf{.047} \\
ELAM / Llama   & \textbf{$-$22.4} & \textbf{9.0e$-$11} & \textbf{$-$36.0} & \textbf{4.8e$-$18} & $+$18.4 & 4.7e$-$12 \\
ELAM / Gemma   & \textbf{$-$29.4} & \textbf{4.0e$-$20} & \textbf{$-$37.8} & \textbf{2.6e$-$25} & $-$1.8 & 1.000 \\
\cmidrule(lr){1-7}
Natural / Qwen  & \textbf{$-$67.8} & \textbf{$<$1e$-$300} & \textbf{$-$69.6} & \textbf{$<$1e$-$300} & \textbf{$-$26.3} & \textbf{3.1e$-$13} \\
Natural / Llama & $+$9.7 & 2.0e$-$4 & \textbf{$-$32.2} & \textbf{6.9e$-$128} & $+$38.7 & 1.5e$-$155 \\
Natural / Gemma & \textbf{$-$60.3} & \textbf{$<$1e$-$300} & \textbf{$-$49.8} & \textbf{2.0e$-$257} & $-$7.1 & .389 \\
\bottomrule
\end{tabular}
\end{table*}

\paragraph{Results.}
As reported in Table~\ref{tab:significance_main}, OBJECTION reduces FGR significantly in 20 of the 27 comparisons. Against the single-pass baselines the effect is near-uniform: 9 of 9 against Self-Refine and 8 of 9 against Baseline D, with the sole exception on Natural Innocent with Llama-3.1, where Baseline D already over-acquits (FNR 45.08\%) and therefore starts from an atypically low FGR. Against Debate-Feedback the outcome is backbone-dependent---3 significant wins, 3 losses, and 3 ties---which is consistent with the volatility reported in Section~\ref{sec:analysis}: on Llama-3.1 the debate collapses into indiscriminate acquittal.

\section{Extended Experiments}
\subsection{Coverage across Acquittal Grounds}
\label{sec:coverage}

As shown in the Lawyer Agent prompt (Appendix~\ref{sec:appendix}), the prompt presents a set of legal doctrines appropriate to each acquittal phase. One might object that this is too explicit a hint. We therefore tested the pipeline on acquittal grounds that do not appear in the prompt, and the performance generalized and held up. Table~\ref{tab:coverage_grounds} breaks down FGR by the ground on which the court actually acquitted; the reduction is large and consistent across all three stages, including the Offense stage, which is dominated by evidentiary insufficiency rather than by any named doctrine.

\begin{table}[h]
\centering
\small
\caption{FGR (\%) by acquittal ground on the Natural Innocent dataset (Qwen-2.5-7B).}
\label{tab:coverage_grounds}
\begin{tabular}{lrrr}
\toprule
\textbf{Acquittal Ground} & \textbf{$n$} & \textbf{Baseline D} & \textbf{Ours} \\
\midrule
Offense       & 1,464 & 85.79 & \textbf{17.08} \\
Unlawfulness  &   431 & 86.31 & \textbf{17.40} \\
Culpability   &   226 & 72.57 & \textbf{12.83} \\
\midrule
All           & 2,121 & 84.49 & \textbf{16.69} \\
\bottomrule
\end{tabular}
\end{table}

\subsection{Failure Mode Analysis}
\label{sec:failure_modes}

\begin{table}[h]
\centering
\small
\caption{Distribution of the 404 false acquittals on the Natural Innocent dataset (Qwen-2.5-7B).}
\label{tab:failure_modes}
\begin{tabular}{@{}lrr@{}}
\toprule
\textbf{Breakdown} & \textbf{$n$} & \textbf{\%} \\
\midrule
\multicolumn{3}{@{}l}{\textit{By exit stage}} \\
\quad Offense              & 370 & 91.6 \\
\quad Unlawfulness         &  21 &  5.2 \\
\quad Culpability          &  13 &  3.2 \\
\midrule
\multicolumn{3}{@{}l}{\textit{By charge type}} \\
\quad Assault / bodily injury & 171 & 42.3 \\
\quad Fraud / embezzlement    & 104 & 25.7 \\
\quad Other                   & 129 & 31.9 \\
\bottomrule
\end{tabular}
\end{table}

Table~\ref{tab:failure_modes} breaks down the 404 false acquittals produced on Natural Innocent with Qwen-2.5-7B. The errors concentrate at the Offense stage, and roughly 68\% of those exits show a verdict--reasoning inconsistency---the reasoning finds the elements satisfied while the emitted label is \texttt{NO}---a self-consistency failure of the backbone that is markedly rarer on the commercial APIs (Table~\ref{tab:commercial_analysis}). The errors are also skewed by charge type: assault cases typically involve over-crediting a self-defense narrative in a mutual altercation, and fraud cases an uncorroborated negation of intent, both of which bear on public safety and motivate the human-in-the-loop deployment described in Section~\ref{ethics}.

\section{Licenses and Attribution}
\label{sec:licenses}

\paragraph{External resources.}
Table~\ref{tab:licenses} summarizes the external datasets, models, and software used in this work. All resources are used for research and evaluation purposes only.

\begin{table*}[h]
\centering
\small
\renewcommand{\arraystretch}{1.3}
\begin{tabular}{llll}
\toprule
\textbf{Resource} & \textbf{Type} & \textbf{License} & \textbf{Use in this work} \\
\midrule
\shortstack[l]{CAIL2018 \\ {\scriptsize\cite{xiao2018cail}}}
  & Dataset & MIT & In-domain LJP benchmark. \\
\shortstack[l]{ELAM (via LJPIV) \\ {\scriptsize\cite{yu2022explainable}}}
  & Dataset & MIT & Out-of-domain LJP benchmark. \\
\midrule
\shortstack[l]{Qwen-2.5-7B-Instruct \\ {\scriptsize\cite{qwen2.5}}}
  & LLM & Apache 2.0 & Primary inference backbone. \\
\shortstack[l]{Llama-3.1-8B-Instruct \\ {\scriptsize\cite{dubey2024llama}}}
  & LLM & Llama 3.1 Community & Robustness evaluation backbone. \\
\shortstack[l]{Gemma-2-9B-it \\ {\scriptsize\cite{gemma_2024}}}
  & LLM & Gemma Terms of Use & Robustness evaluation backbone. \\
\midrule
vLLM v0.11.0  & Software & Apache 2.0 & LLM inference server. \\
PEFT v0.17.1  & Software & Apache 2.0 & LoRA fine-tuning. \\
\midrule
\shortstack[l]{Self-Refine \\ {\scriptsize\cite{madaan2023self}}}
  & Baseline & Apache 2.0 & Generic critic baseline. \\
\shortstack[l]{LJPIV \\ {\scriptsize\cite{zhang2025ljpiv}}}
  & Baseline & Apache 2.0 & Fine-tuned SOTA baseline. \\
\shortstack[l]{Debate-Feedback \\ {\scriptsize\cite{chen2025debate}}}
  & Baseline & Apache 2.0 & Multi-agent debate baseline. \\
\bottomrule
\end{tabular}
\caption{Licenses and intended use of external resources. All are used solely for research purposes.}
\label{tab:licenses}
\end{table*}

\end{document}